\documentclass{article}
\PassOptionsToPackage{numbers,sort&compress}{natbib}
\usepackage[main, final]{neurips_2026}

\usepackage{tabularx}
\usepackage{booktabs}
\usepackage{array}
\newcolumntype{L}[1]{>{\raggedright\arraybackslash}p{#1}}
\newcolumntype{Y}{>{\raggedright\arraybackslash}X}

\usepackage{etoc}
\usepackage{float}
\usepackage[utf8]{inputenc}
\usepackage[T1]{fontenc}
\usepackage{hyperref}
\usepackage{url}
\usepackage{booktabs}
\usepackage{amsfonts}
\usepackage{amsmath}
\usepackage{amssymb}
\usepackage{amsthm}
\usepackage{bm}
\usepackage{nicefrac}
\usepackage{microtype}
\usepackage[table]{xcolor}
\usepackage{graphicx}
\usepackage{multirow}
\definecolor{myorange}{RGB}{220,110,0}
\hypersetup{
    colorlinks=true,
    linkcolor=myorange,
    citecolor=myorange,
    urlcolor=myorange
}

\title{The Locality Trade-off in Masked Patch Self-Distillation}

\author{
    Xi Weng$^{1}$, Zhaoyu Liu$^{1}$,  Lianyu Hu$^{2}$, Jin Song Dong$^{1}$\thanks{corresponding author: dongjs@comp.nus.edu.sg}
    \\
 $^{1}$School of Computing, National University of Singapore\\
 $^{2}$Nanyang Technological University\\
}

\newcommand{\LDINO}{\mathcal{L}_{\mathrm{DINO}}}

\newcommand{\LIBOT}{\mathcal{L}_{\mathrm{iBOT}}}

\newcommand{\sg}{\operatorname{sg}}

\definecolor{semanticbg}{HTML}{E8F5E9}
\definecolor{localitybg}{HTML}{FDE7E7}
\definecolor{improvegreen}{HTML}{2A7A2A}
\definecolor{degradered}{HTML}{A02020}

\begin{document}

\maketitle
\begin{abstract}
Masked patch self-distillation is a key component of modern vision
foundation models, including DINOv2 and DINOv3, yet its implications
for local visual information remain unclear. We show that its semantic
benefits can coexist with a scale-dependent locality cost. Using two
annotation-free diagnostics, we distinguish recoverable local appearance
from spatial correspondence. In smaller models, adding iBOT substantially
weakens patch correspondence while largely preserving appearance
decodability. In larger models, correspondence improves, but pixel and
texture evidence becomes less linearly accessible in deep features.
Thus, scaling changes the form of the locality cost rather than
uniformly eliminating it. Our analysis of Gram anchoring further shows
that preserving patch relations does not guarantee simultaneous
improvements in both properties. These findings motivate evaluating
and designing visual representations around the complementary demands
of semantic abstraction, appearance preservation, and spatial
correspondence. Our code is available at \texttt{https://github.com/winci-ai/jea}
\end{abstract}

\vspace{-0.15in}
\section{Introduction}
\label{sec:introduction}

\begin{figure}[H]
  \centering
  \includegraphics[width=14cm]{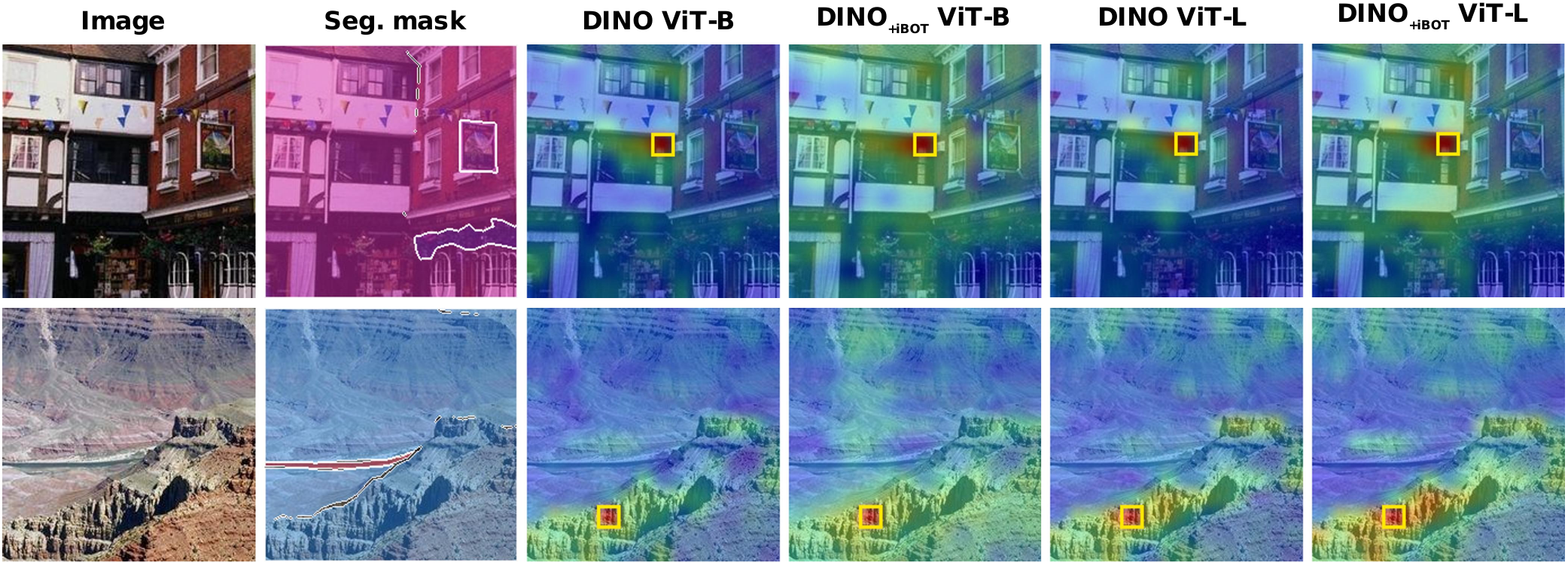}
  \vspace{-0.2in}
  \caption{ Anchor-patch similarity heatmaps for DINO and DINO$_{+\mathrm{iBOT}}$ on ADE20K examples, with semantic annotations shown for reference. Yellow boxes mark the anchor
  patches. For ViT-B, adding iBOT produces broader responses over the building and mountain, reducing distinctions associated with windows and rugged cliff structures. This effect is alleviated in ViT-L, where the $+$iBOT model also exhibits more
  spatially selective responses to these structures. The semantic masks do not resolve many of these within-region
  differences, illustrating why segmentation performance alone cannot characterize fine-grained local representations.
  Additional examples appear in Figure~\ref{fig:heatmap_extra}.}
  \label{fig:teaser}
\end{figure}

Self-supervised learning has become a central approach to building
general-purpose visual representations~\citep{2020_CVPR_He,2020_ICML_Chen,caron2021dino,chen2021mocov3,2020_NIPS_Grill,2020_NIPS_Caron,2022_ICLR_Adrien,2022_NIPS_Weng,2024_ICLR_Weng,2024_TPAMI_Huang,weng2025resa}. From DINO~\cite{caron2021dino}
to DINOv2~\cite{oquab2023dinov2} and
DINOv3~\cite{simeoni2025dinov3}, learning from unlabeled images has
produced visual backbones that transfer across recognition, dense
prediction, and geometric understanding. Their influence also extends
to larger systems: DINOv2 contributes visual features to multimodal
models such as Cambrian-1~\cite{tong2024cambrian} and serves as the
backbone of Depth Anything V2~\cite{yang2024depth}.
As these representations become shared components across applications,
understanding which visual properties their training objectives
preserve becomes increasingly important.

A key ingredient in this development is masked patch
self-distillation. Introduced by iBOT~\cite{zhou2022ibot} and adopted
in DINOv2 and DINOv3, this objective trains masked student patches
to predict teacher-generated targets. It extends self-distillation
to patch tokens, enabling supervision at a finer spatial granularity.
However, a patch-level objective does not necessarily preserve all
forms of local evidence. Because its targets are learned
representations rather than raw pixels, successful prediction may
favor semantic consistency while becoming less sensitive to
appearance differences within a semantic region. This raises a
fundamental question: \emph{what does masked patch self-distillation
preserve locally, and how does the answer change with model scale?}

Figure~\ref{fig:teaser} illustrates this question using images from
ADE20K~\cite{zhou2017scene}. For ViT-B, DINO produces similarity
responses that distinguish structures such as windows and rugged
cliffs, whereas adding iBOT yields broader responses over the
surrounding building or mountain. These differences can be obscured
by semantic segmentation evaluation: human-annotated masks need not
distinguish the visual structures contained within a labeled region.
Interestingly, this qualitative loss of spatial specificity is
alleviated in ViT-L, where the $+$iBOT model also exhibits responses
that follow fine local structures. Yet sharper similarity maps alone
cannot establish whether the underlying pixel and texture evidence
remains recoverable.

To investigate this distinction, we compare DINO and
DINO$_{+\mathrm{iBOT}}$ across four backbone scales under matched
training settings within each scale. The downstream results reveal
a consistent semantic advantage alongside a more heterogeneous
effect on local tasks. Adding iBOT generally improves tasks dominated
by semantic understanding, while often reducing performance on tasks
that require precise local distinctions. However, this pattern is
not universal: on DAVIS video object tracking, the disadvantage
observed in smaller models disappears and reverses slightly in
larger models. Together with Figure~\ref{fig:teaser}, this suggests
that a single notion of ``local information'' is insufficient to
explain the effects of patch self-distillation.

We therefore introduce two complementary annotation-free diagnostics.
\emph{Local Appearance Decodability} measures how much pixel and
texture evidence is linearly recoverable from frozen patch tokens.
\emph{Local-to-Global Correspondence} measures whether patches remain
spatially distinguishable and matchable across views.
These diagnostics reveal a scale-dependent dissociation.
For ViT-S and ViT-B, iBOT substantially weakens correspondence while
leaving appearance decodability largely unchanged or slightly improved.
For ViT-L and ViT-H, correspondence is preserved or enhanced, but
appearance decodability declines markedly in deep layers.
The improved spatial specificity of larger models therefore does
not imply that all forms of local evidence are preserved.
Our results instead suggest that model scale changes which local
properties are most affected by semantic patch supervision.

Finally, we examine Gram anchoring, introduced in
DINOv3~\cite{simeoni2025dinov3} to stabilize dense features by
regularizing pairwise patch similarities. We compare anchoring
initialized from an earlier DINO$_{+\mathrm{iBOT}}$ checkpoint with
a variant initialized from DINO. The resulting changes again
separate appearance from correspondence: the DINO-based anchor
substantially improves correspondence in ViT-B, whereas in ViT-L
it improves deep appearance decodability without consistently
improving correspondence. These results suggest that the local
properties encouraged by relational regularization depend on both
the anchor representations and the model scale. 

Our contributions are threefold:
\begin{itemize}
    \item We characterize the scale-dependent locality cost of
    masked patch self-distillation, showing that its semantic
    benefits coexist with distinct changes in local visual abilities.
    \item We introduce two annotation-free diagnostics that separate
    local appearance decodability from spatial correspondence,
    revealing differences that downstream semantic evaluation and
    similarity visualizations alone cannot resolve.
    \item We analyze Gram anchoring through this distinction,
    showing that improvements in patch relations and recoverable
    appearance need not occur together.
\end{itemize}

\section{Background, Related Work, and Notation}
\label{sec:methods}
\vspace{-0.1in}
In this section, we provide a detailed description of the class-token joint-embedding architecture and masked patch self-distillation. Additional related work is discussed in Appendix~\ref{app:related_work}.

\subsection{Class-Token Joint-Embedding Self-Distillation}
\label{sec:cls_self_distillation}

Let $\mathcal B=\{x_i\}_{i=1}^{B}$ be a mini-batch of unlabeled images
sampled from the data distribution $\mathcal D$. For each image $x_i$,
a multi-crop augmentation pipeline~\cite{2020_NIPS_Caron} generates
global and local views, denoted by $\mathcal V_i^{g}$ and
$\mathcal V_i^{\ell}$, respectively. Following
DINO~\cite{caron2021dino}, the student processes all views, whereas
the teacher processes only the global views. We denote their respective
crop sets by
\begin{equation}
  \mathcal C_i
  =
  \mathcal V_i^{g}\cup\mathcal V_i^{\ell},
  \qquad
  \mathcal G_i
  =
  \mathcal V_i^{g}.
\end{equation}

Given a crop $w$, a vision transformer
encoder~\cite{dosovitskiy2020image} produces a class token and a
sequence of patch tokens:
\begin{equation}
  f_{\theta}(w)
  =
  \big(
    c_{\theta}(w),
    z_{\theta}(w,1),
    \ldots,
    z_{\theta}(w,P_w)
  \big),
\end{equation}
where $c_{\theta}(w)$ denotes the class-token representation,
$z_{\theta}(w,r)$ denotes the $r$-th patch-token representation,
and $P_w$ is the number of patch tokens in $w$.
Let $a_i(u)$ denote the student input for crop $u$: it applies the
sampled patch mask to global crops when iBOT is enabled and is the
identity otherwise. Local crops and teacher inputs remain unmasked.
The class-token representations are mapped to $K$-dimensional
logit vectors by the student and teacher projection heads:
\begin{equation}
  \begin{aligned}
    \bm s_i^{u}
    =
    g_{\theta}^{\mathrm{cls}}
    \big(c_{\theta}(a_i(u))\big),
     u\in\mathcal C_i,
    &&\bm t_i^{v}
    =
    g_{\bar\theta}^{\mathrm{cls}}
    \big(c_{\bar\theta}(v)\big),
     v\in\mathcal G_i.
  \end{aligned}
\end{equation}
Here, $\theta$ and $\bar\theta$ collectively denote the parameters
of the student and teacher encoders and projection heads,
respectively. The teacher is initialized from the student and
receives no gradient updates. Instead, after each student update,
its parameters are updated by an exponential moving average:
\begin{equation}
  \bar\theta
  \leftarrow
  \mu\bar\theta+(1-\mu)\theta,
\end{equation}
where $\mu\in[0,1)$ is a momentum coefficient, typically increased
toward $1$ during training.

We use DINO~\cite{caron2021dino} as the class-token
self-distillation objective. DINO encourages agreement between
different views of the same image by matching student predictions
to stop-gradient teacher targets. The original method uses
centering and sharpening of teacher outputs to help avoid
collapse. To control for differences in the class-token objective when
comparing DINO with DINO$_{+\mathrm{iBOT}}$, we instead construct teacher targets
using Sinkhorn--Knopp normalization, following
SwAV~\cite{2020_NIPS_Caron} and
DINOv2~\cite{oquab2023dinov2}, while retaining
temperature-controlled sharpening.
Details of the Sinkhorn--Knopp target construction are provided in
Appendix~\ref{sec:cls_objectives}.

\vspace{-0.05in}
\subsection{Masked Patch Self-Distillation: iBOT}
\label{sec:ibot_loss}
\vspace{-0.05in}

iBOT~\citep{zhou2022ibot} complements class-token
self-distillation with a masked patch prediction objective.
The student predicts teacher-generated target distributions
at masked patch positions, using the corresponding unmasked
crop as the teacher input. Thus, the prediction targets
are learned patch-level distributions rather than raw pixels.

For each global crop $u\in\mathcal G_i$, let
$\mathcal M_i^{u}\subseteq\{1,\ldots,P_u\}$ denote the
selected masked patch indices. We write $\widetilde u$
for the student input obtained by replacing the selected
patch embeddings with a learnable mask token, so that
$a_i(u)=\widetilde u$. The class and patch losses use the same
masked student forward pass.
The teacher receives the unmasked crop $u$.
At each masked position $r\in\mathcal M_i^{u}$,
the student and teacher patch heads produce logits
\begin{equation}
  \begin{aligned}
    \bm a_{i,r}^{s,u}
    &=
    h_{\theta}^{\mathrm{patch}}
    \big(z_{\theta}(\widetilde u,r)\big),
    &\bm a_{i,r}^{t,u}
    &=
    h_{\bar\theta}^{\mathrm{patch}}
    \big(z_{\bar\theta}(u,r)\big),
  \end{aligned}
\end{equation}
where both logit vectors belong to $\mathbb R^{K_p}$,
and $K_p$ is the output dimension of the patch heads.
Since the student and teacher inputs share the same
crop geometry, the index $r$ identifies the same spatial
position in both networks.

The student patch prediction is
  $\bm p_{i,r}^{s,u}
  =
  \operatorname{softmax}
  \left(
    \frac{\bm a_{i,r}^{s,u}}{\tau_s^{p}}
  \right)$, where $\tau_s^{p}>0$ is the student patch temperature.
Following DINOv2~\citep{oquab2023dinov2}, we construct
teacher patch targets using Sinkhorn--Knopp normalization.
Specifically, let $\bm A_{\mathcal B}^{t}$ stack the
teacher patch logits at all selected masked positions
in the mini-batch. The corresponding target matrix is
  $\bm Q_{\mathcal B}^{t}
  =
  \sg\!\left[
    \operatorname{SK}
    \left(
      \exp\left(
        \frac{\bm A_{\mathcal B}^{t}}{\tau_t^{p}}
      \right)
    \right)
  \right]$, where $\tau_t^{p}>0$ is the teacher patch temperature,
the exponential is applied elementwise, and
$\operatorname{SK}$ uses the same normalization convention
as in Appendix~\ref{sec:cls_objectives}.
Each row provides the target distribution
$\bm q_{i,r}^{t,u}$ for its corresponding patch.
Normalization is performed jointly across the selected
teacher patches, and all teacher targets are detached
from gradient computation.

Defining
$H(\bm q,\bm p)=-\sum_{k=1}^{K_p}q_k\log p_k$,
the masked patch self-distillation loss is
\begin{equation}
  \LIBOT
  =
  \frac{1}{B}
  \sum_{i=1}^{B}
  \frac{1}{|\mathcal G_i|}
  \sum_{u\in\mathcal G_i}
  \frac{
    \displaystyle
    \sum_{r\in\mathcal M_i^{u}}
    H\!\left(
      \bm q_{i,r}^{t,u},
      \bm p_{i,r}^{s,u}
    \right)
  }{
    \max\big\{1,|\mathcal M_i^{u}|\big\}
  }.
  \label{eq:ibot_ce_loss}
\end{equation}
The loss is evaluated only at masked positions and
averaged over global crops and images.
Crops with no masked positions contribute zero.
The student encoder and patch head are optimized by
back-propagation, while their teacher counterparts
are updated by exponential moving average.

\begin{figure}[t]
  \centering
  \makebox[\textwidth][c]{%
    \includegraphics[width=14cm]
{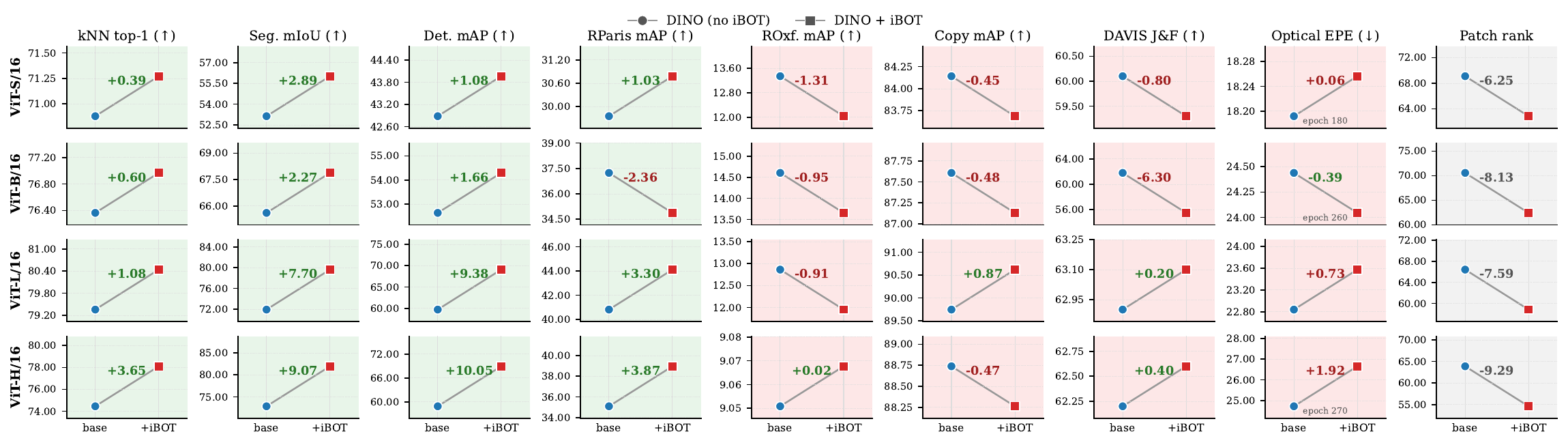}
}
  \caption{Downstream comparison of DINO and DINO$_{+\mathrm{iBOT}}$ across
ViT-S/16, ViT-B/16, ViT-L/16, and ViT-H/16, pretrained on
ImageNet-1K for 300 epochs under matched settings within each scale.
Adding iBOT generally benefits tasks dominated by semantic
understanding, whereas its effects on tasks requiring fine-grained
local information vary with model scale. In particular, the
degradation in DAVIS tracking observed for smaller backbones
reverses to modest gains for larger backbones. Task scores other than optical-flow EPE are percentages;
normalized patch rank is multiplied by 100.
Arrows indicate the preferred direction for task metrics:
$\uparrow$ denotes higher is better and $\downarrow$ denotes
lower is better. Patch effective rank measures spectral diversity.
Numerical results and training trajectories are provided in
Table~\ref{tab:fig2_vit_base_values} and
Figure~\ref{fig:trajectories}, respectively.
}
  \label{fig:paired_delta}
  \vspace{-0.15in}
\end{figure}

\vspace{-0.1in}
\section{The Downstream Signature of iBOT's Locality Cost}
\label{sec:phenomenon}
\vspace{-0.075in}

In this section, we compare DINO and DINO$_{+\mathrm{iBOT}}$ on ViT-S/16, ViT-B/16,
ViT-L/16, and ViT-H/16, all pretrained on ImageNet-1K~\cite{2009_ImageNet} for 300 epochs
following the DINOv2 recipe~\cite{oquab2023dinov2}. Within each scale,
we keep the remaining training settings identical and enable or disable
the iBOT branch together with its associated student masking.

Our evaluations cover complementary representation capabilities.
ImageNet $k$-NN classification~\cite{caron2021dino}
primarily measures global semantic discrimination;
VOC2012 segmentation~\cite{everingham2015pascal} and
VOC2007 detection~\cite{2010_IJCV_VOC} require semantic recognition
together with spatial localization.
RParis and ROxford retrieval~\cite{radenovic2018revisiting} evaluate
instance discrimination using both global structure and distinctive
local cues, while Copydays copy detection~\cite{douze2009evaluation}
emphasizes visual identity under image transformations.
DAVIS label propagation~\cite{ponttuset2017davis,caron2021dino}
tests cross-frame patch matching and instance consistency, and
Sintel optical flow~\cite{butler2012naturalistic} probes motion
estimation through a learned head. Our Sintel scores measure fitting
on training videos, not held-out generalization.
We additionally report patch effective rank
as a measure of spectral diversity
\cite{roy2007effective,garrido2023rankme}.
Benchmark interpretations and evaluation details are provided in
Appendix~\ref{app:benchmark_capabilities} and
Appendix~\ref{app:experimental_settings}.

\vspace{-0.025in}
Figure~\ref{fig:paired_delta} reveals a recurring contrast:
DINO$_{+\mathrm{iBOT}}$ improves classification, segmentation, and
detection across all scales, whereas DINO more often performs better
on tasks emphasizing local visual distinctiveness, including Copydays
and ROxford retrieval. RParis generally benefits from iBOT, motivating
separate reporting of the two retrieval benchmarks.
However, the locality-sensitive results are not uniform. Most notably,
iBOT reduces DAVIS performance on ViT-S/B but modestly improves it on
ViT-L/H. Since our DAVIS evaluation relies on patch similarities for
label propagation, this exception suggests that iBOT may improve
patch correspondence in larger models. Patch effective rank decreases
at every scale, showing that reduced spectral diversity need not
imply weaker matching.

\begin{figure}[t]
  \centering
  \includegraphics[width=11cm]{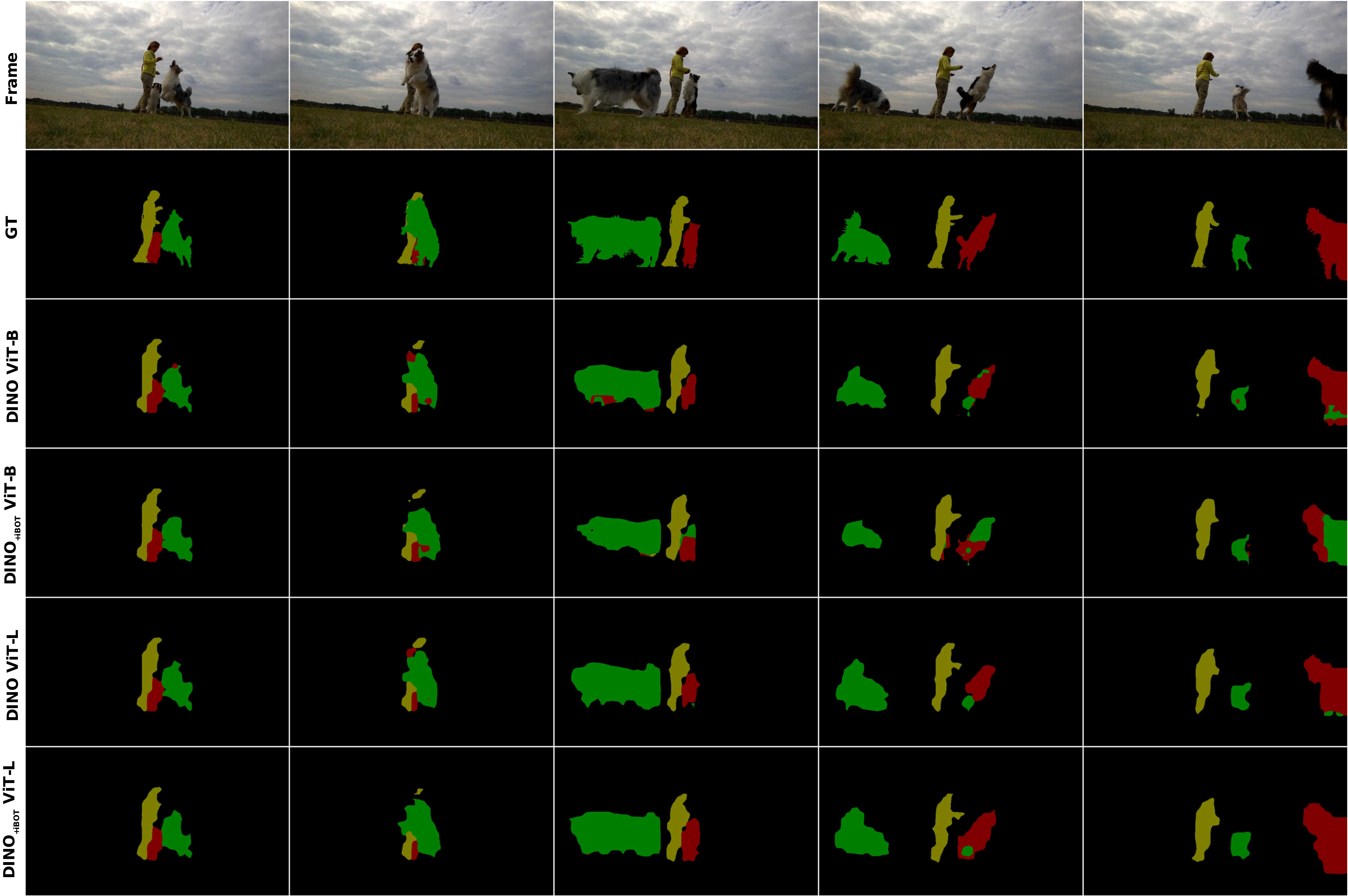}
  \vspace{-0.05in}
  \caption{DAVIS video object tracking visualization. Here we compare tracking masks produced by DINO and DINO$_{+\mathrm{iBOT}}$ on videos containing visually similar same-category instances and occlusions. More visualization results can be seen in Figure~\ref{fig:davis_extra}.}
  \label{fig:davis_visual}
   \vspace{-0.15in}
\end{figure}

Figure~\ref{fig:davis_visual} further visualizes DAVIS tracking
predictions from DINO and DINO$_{+\mathrm{iBOT}}$. With ViT-B/16,
the $+$iBOT model exhibits two representative failure modes.
First, when visually similar instances of the same category,
such as two dogs, overlap, it is more prone to identity confusion
and swapped tracking masks. Second, after an instance becomes
occluded, it is less successful at recovering and maintaining
the correct track. These failures are alleviated when scaling
to ViT-L/16, consistent with the scale-dependent quantitative
results. This qualitative evidence further motivates examining
how iBOT affects instance-specific matching across model scales.

Overall, the downstream results indicate a recurring locality cost
whose form depends on the task and model scale. They cannot determine
whether the regressions reflect reduced appearance information,
less effective matching geometry, or both. We therefore introduce
two complementary annotation-free diagnostics:
\emph{Local Appearance Decodability} measures recoverable pixel and
texture evidence, whereas \emph{Local-to-Global Correspondence}
measures whether patches remain spatially distinguishable and
matchable across views.

\begin{figure}[t]
  \centering
  \makebox[\textwidth][c]{%
    \includegraphics[width=13cm]{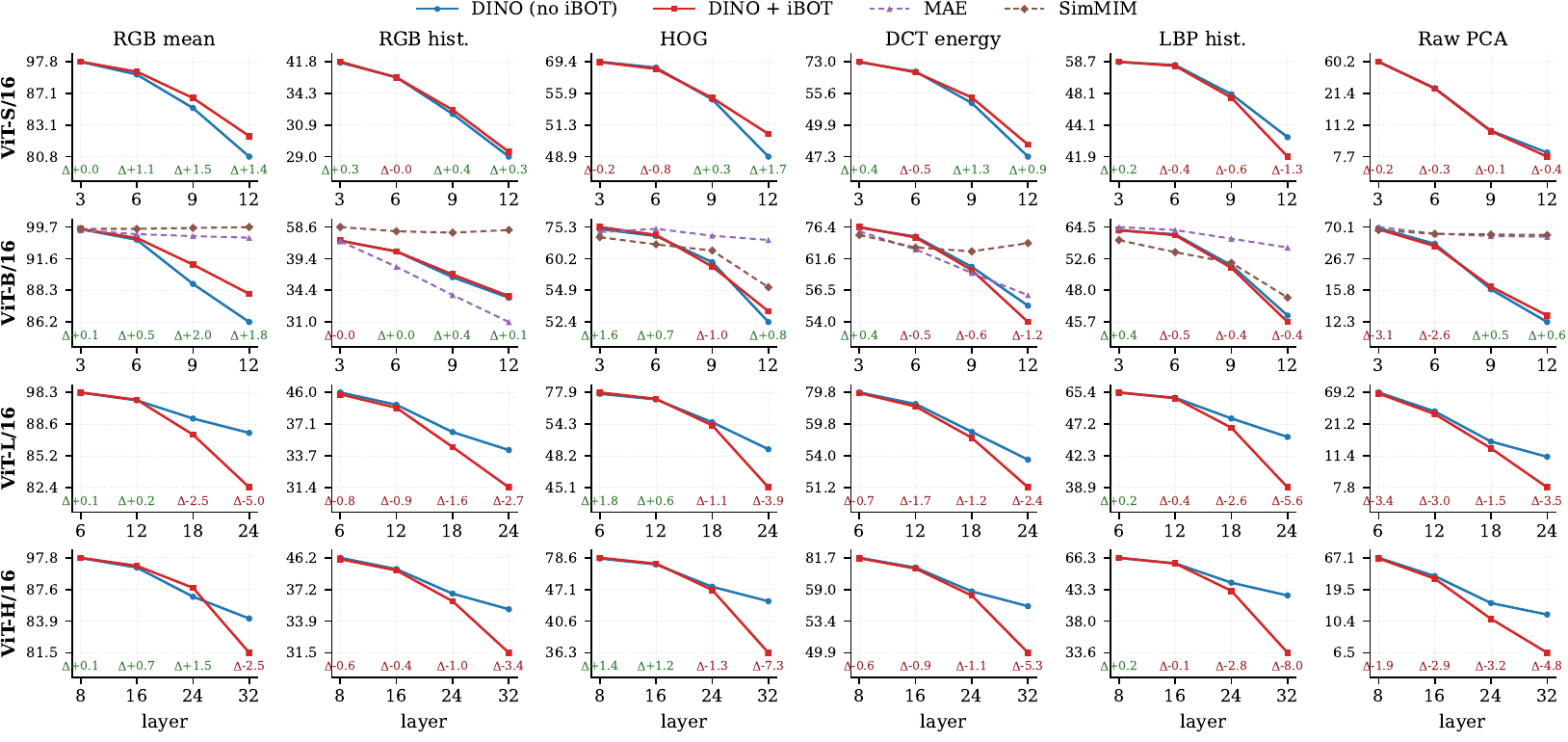} }
    \vspace{-0.25in}
  \caption{Quantitative results of Local appearance decodability across different layers. $\Delta$ denotes the difference between the DINO$_{+\mathrm{iBOT}}$ and its corresponding DINO model under the same pretraining setup. To validate that the diagnostic indeed captures local visual evidence, we also include raw-pixel prediction models, such as SimMIM and MAE, which are also pretrained using ViT-B/16 on ImageNet. We also report the evolution of each descriptor metric during training in Figure~\ref{fig:decodability_trajectory}, and show the decoding visualization in Figure~\ref{fig:local_decode_visual}. }
  \label{fig:decodability_layers}
    \vspace{-0.25in}
\end{figure}

\vspace{-0.15in}
\section{Locating Lost Evidence via Annotation-Free Diagnostics}
\label{sec:mechanism}
\vspace{-0.1in}
The downstream results in Section~\ref{sec:phenomenon} reveal
locality-sensitive regressions alongside scale-dependent exceptions,
particularly the improved DAVIS performance of larger models.
However, these benchmarks combine semantic recognition, appearance
discrimination, and spatial matching, making it difficult to isolate
which properties change under iBOT supervision. We therefore introduce
two complementary annotation-free diagnostics.
\emph{Local Appearance Decodability} measures recoverable pixel and
texture evidence, whereas \emph{Local-to-Global Correspondence}
measures whether patches remain spatially distinguishable and
matchable across views. This distinction allows us to examine whether
the downstream locality cost reflects changes in appearance
accessibility, correspondence, or both.

\vspace{-0.15in}
\subsection{Local Appearance Decodability}
\label{sec:decodability}
\vspace{-0.1in}
The first diagnostic measures how much local appearance information
can be linearly recovered from frozen patch tokens.
Given an image $w\in[0,1]^{3\times H\times H}$, let
$z_{\theta}^{(\ell)}(w,r)\in\mathbb{R}^{d}$ denote its $r$-th patch
token at layer $\ell$, and let
$\tilde w_r\in[0,1]^{3\times p\times p}$ be the spatially aligned
RGB patch, where $p$ is the patch size.

We construct an appearance target
$y(w,r)=\phi(w,r)\in\mathbb{R}^{m}$ from RGB mean, RGB
histograms, Histogram of Oriented Gradients (HOG), Discrete Cosine
Transform (DCT) frequency-band energy, Local Binary Patterns (LBP)
histograms, and a low-dimensional raw-pixel PCA projection.
These descriptors capture complementary aspects of color, edge
orientation, texture, spatial frequency, and pixel variation without
semantic annotations. HOG and LBP use image-level neighborhood
operations followed by patch-wise pooling; the other descriptors
are computed within each aligned raw patch.

We fit a ridge linear decoder from frozen patch tokens to these
targets using a probe-training split and evaluate it on held-out
images. For each descriptor group $g$, we report the
dimension-averaged coefficient of determination,
$R^2_{g,\mathrm{mean}}$. Higher values indicate greater linear
accessibility of the corresponding appearance attributes.
Descriptor construction, probe fitting, and evaluation details are
provided in Appendix~\ref{app:descriptor_decodability}.

We evaluate the 300-epoch backbones from
Section~\ref{sec:phenomenon} at multiple depths, since dense
prediction commonly uses intermediate as well as final-layer
features~\cite{caron2021dino,zhou2022ibot}.
We sample blocks 3/6/9/12 for ViT-S/B, 6/12/18/24 for ViT-L,
and 8/16/24/32 for ViT-H.
Figure~\ref{fig:decodability_layers} additionally includes
ViT-B/16 MAE~\cite{2022_cvpr_He} and
SimMIM~\cite{xie2022simmim} as pixel-reconstruction references.
Their strong deep-layer RGB-mean and raw-PCA scores provide
supporting evidence that the diagnostic captures appearance
information beyond semantic discrimination.

Figure~\ref{fig:decodability_layers} reveals a clear dependence
on model scale and depth.
For ViT-S/B, adding iBOT largely preserves appearance decodability,
with modest gains or losses depending on the descriptor and layer.
In particular, final-layer RGB-mean and HOG scores improve for
both backbones. Thus, the smaller models do not exhibit a systematic
loss of linearly recoverable appearance information.
Their weaker performance on locality-sensitive downstream tasks,
including DAVIS, therefore cannot be explained simply by a broad
decline in the appearance attributes measured here.

For ViT-L/H, iBOT provides small improvements in selected shallow
or intermediate-layer descriptors, including HOG and RGB mean.
However, this benefit does not extend to the final representations:
all six descriptor scores decrease at the last block, with
particularly pronounced losses in HOG and LBP for ViT-H.
The appearance cost is therefore concentrated in deeper
representations, rather than occurring uniformly throughout the
backbone. These results indicate that iBOT makes pixel and texture
evidence less linearly accessible in the deep layers of larger models.

Together with the downstream results, this pattern highlights why
appearance decodability and correspondence must be examined separately.
ViT-S/B retain broadly comparable appearance decodability despite
worse DAVIS tracking, whereas ViT-L/H improve on DAVIS despite
reduced deep-layer appearance decodability.
This dissociation suggests that recoverable appearance alone does
not determine matching quality. We next directly evaluate patch
correspondence to test whether its response to iBOT also changes
with model scale.

\begin{figure}[t]
  \centering
  \includegraphics[width=11cm]{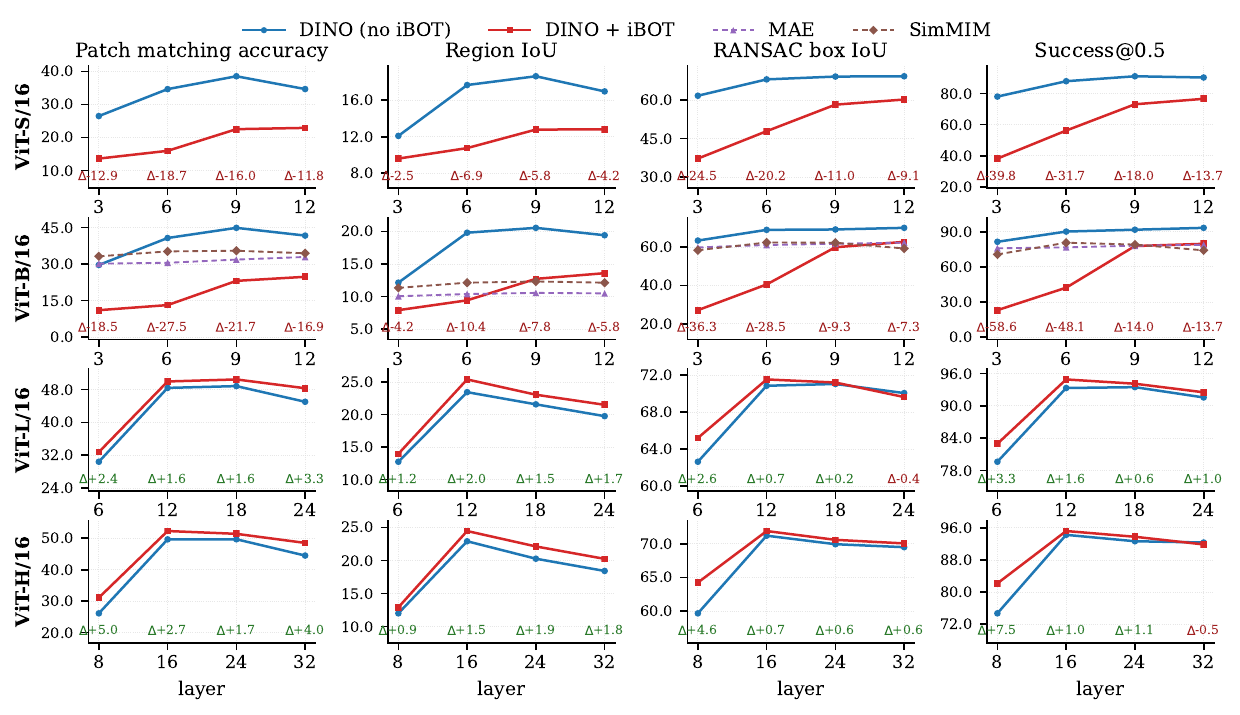}
  \vspace{-0.15in}
  \caption{Quantitative results of local-to-global correspondence across different layers. $\Delta$ denotes the difference between the DINO$_{+\mathrm{iBOT}}$ model and its corresponding DINO model under the same pretraining setup. We also report the evolution of each metric during training in Figure~\ref{fig:crop_copy_trajectory}.}
  \label{fig:crop_copy_layers}
  \vspace{-0.15in}
\end{figure}

\vspace{-0.1in}
\subsection{Local-to-Global Correspondence}
\label{sec:diag_local_global}

The second diagnostic evaluates whether patches remain spatially
distinguishable and matchable across views. Given a source image
$x$ and a sampled crop box $B^\star$, we construct a local query
$x_{B^\star}$ by cropping the image inside $B^\star$. Both the
query and the source image are resized to the model input resolution,
and patch tokens are extracted from a frozen layer $\ell$.

Let $\tilde z^{\mathrm{local}}_\ell(i)$ and
$\tilde z^{\mathrm{global}}_\ell(j)$ denote cosine-normalized patch
tokens from the query and source image, respectively. Their matching
scores are
$
A_\ell(i,j)=
\tilde z^{\mathrm{local}}_\ell(i)^\top
\tilde z^{\mathrm{global}}_\ell(j).
$
The ground-truth source region consists of source patches whose grid
centers lie inside $B^\star$.

We report four complementary metrics:
\textbf{patch matching accuracy}, measuring whether query patches
retrieve source patches inside $B^\star$;
\textbf{region localization IoU}, measuring the overlap between the
strongest source responses and the true crop support;
\textbf{RANSAC geometric box IoU}, measuring the accuracy of the crop
box estimated by RANSAC-style translation consensus with known crop
dimensions; and
\textbf{IoU$_{50}$ success rate}, measuring the fraction of estimated
boxes reaching at least $0.5$ IoU.
Higher values indicate better performance under each criterion.
Patch matching accuracy tests region membership, not exact pointwise
correspondence; region IoU also uses the known reference-region size.
Full definitions and sampling details are provided in
Appendix~\ref{app:local_global_details}.

Unlike copy detection, which retrieves an image from a database,
this diagnostic assumes that the source image is known and asks
whether patch similarities can localize the query within it.
It therefore isolates spatial matching more directly, without
image-level aggregation or database distractors.

Figure~\ref{fig:crop_copy_layers} reveals a clear scale-dependent
effect at epoch 300. For ViT-S/B, adding iBOT substantially reduces
all four correspondence metrics across the sampled layers.
At the final block and crop area ratio $0.1$, patch matching accuracy
drops by approximately $11.8$ and $16.9$ percentage points,
respectively. For ViT-L/H, this broad degradation disappears:
patch matching accuracy and region IoU improve across all sampled
layers, with final-block matching gains of approximately $3.3$
and $4.0$ points. Geometric metrics generally improve or remain
comparable, although small exceptions remain, including final-block
box IoU for ViT-L and IoU$_{50}$ success for ViT-H.
Thus, the larger-model benefit is clearest in patch matching and
region localization.

The training trajectories in Figure~\ref{fig:crop_copy_trajectory}
show that these differences develop during optimization.
In ViT-S/B, several $+$iBOT metrics decline over extended periods
of training, particularly toward the end, while DINO generally
improves or maintains its correspondence performance.
At deeper layers, an early $+$iBOT advantage can consequently
turn into a substantial final-checkpoint deficit.
These trajectories are not uniformly monotonic, but consistently
reveal weaker late-training correspondence in the smaller
$+$iBOT models. In ViT-L/H, the sustained deterioration is absent:
correspondence generally improves or stabilizes, and the
$+$iBOT models retain competitive or stronger performance.
This scale dependence agrees with the DAVIS results in
Section~\ref{sec:phenomenon}, where iBOT hurts tracking on ViT-S/B
but modestly improves it on ViT-L/H.

\begin{figure}[t]
  \centering
  \includegraphics[width=11cm]{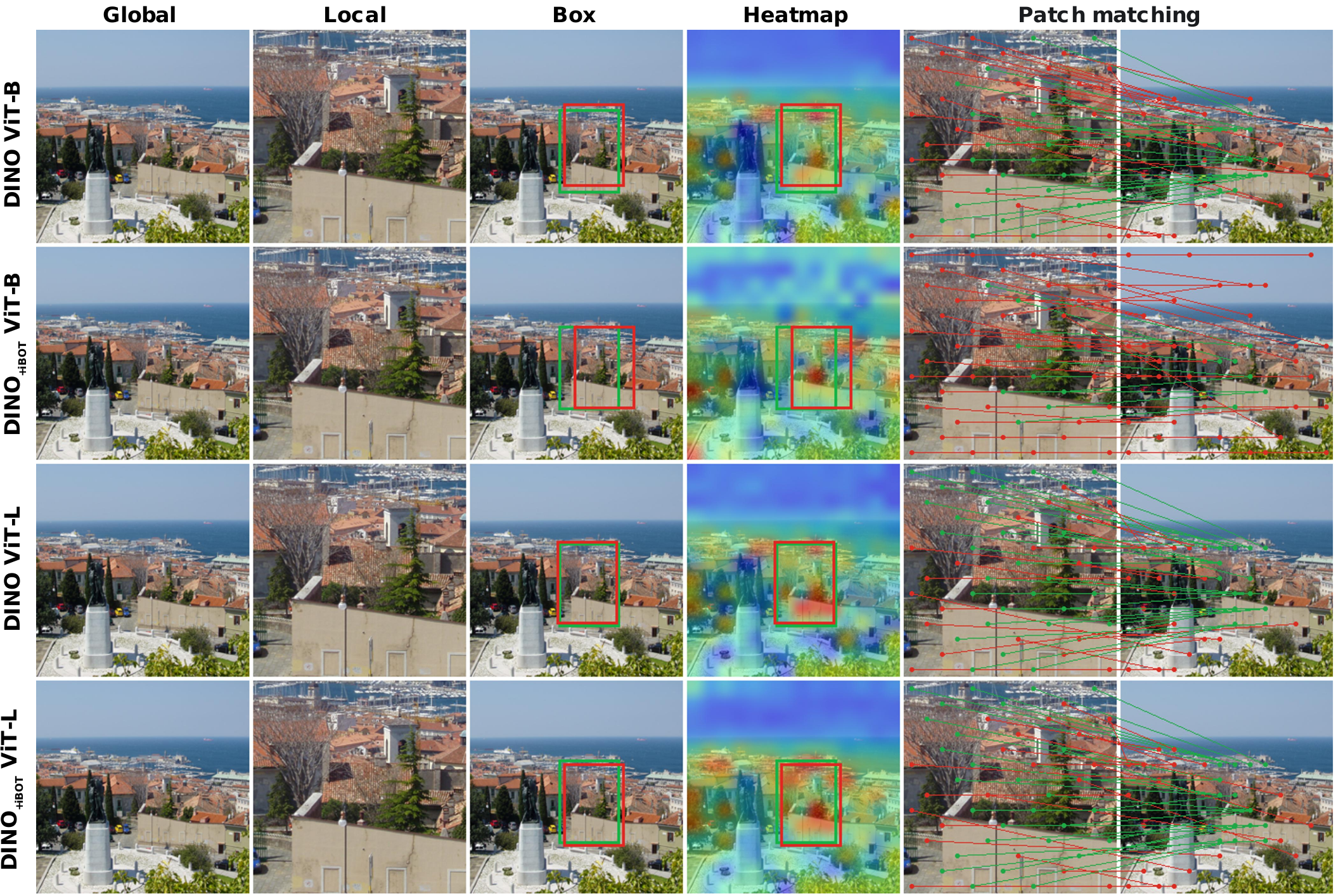}
  \vspace{-0.1in}
 \caption{Local-to-global correspondence visualizations for DINO
and DINO$_{+\mathrm{iBOT}}$, showing RANSAC box estimates,
similarity heatmaps, and patch matches. With ViT-B/16, adding
iBOT weakens correspondence: reliable matches are largely
confined to larger foreground objects, while smaller objects
are incorrectly matched or poorly localized. This failure
pattern is not observed in the ViT-L/16 examples, where the
$+$iBOT model maintains correspondence for smaller objects.
These illustrate the scale-dependent effect reported
in Figure~\ref{fig:crop_copy_layers}. Additional examples appear
in Figure~\ref{fig:local_global_extra}.}
  \label{fig:local_global}
  \vspace{-0.25in}
\end{figure}

Figure~\ref{fig:crop_copy_ratio} examines different crop area ratios.
For ViT-S/B, deficits in patch matching, region IoU, and geometric
box IoU persist across the evaluated ratios, while the
IoU$_{50}$ success gap largely closes for larger crops.
For ViT-L/H, region IoU remains higher with iBOT, whereas the
patch-matching advantage generally narrows as crop size increases
and slightly reverses for ViT-L at ratio $0.5$.
The effect therefore depends on both model scale and the matching
criterion, rather than following a universal relationship with
crop size. Figure~\ref{fig:local_global} provides qualitative
examples of the correspondence failures that the diagnostic captures.

Together with Local Appearance Decodability, these results establish
two distinct responses to iBOT. Smaller models retain broadly
comparable appearance decodability while losing spatial matching
ability. Larger models improve patch matching and region localization
despite reduced deep-layer appearance decodability.
Thus, preserving decodable pixel and texture evidence is insufficient
to guarantee effective correspondence, and better correspondence
does not require preserving all measured appearance attributes.
The locality cost of iBOT consequently takes different forms across
model scales.

\vspace{-0.15in}
\section{How Gram Anchoring Changes Local Representations}
\label{sec:gram}
\vspace{-0.1in}

Section~\ref{sec:mechanism} shows that iBOT affects local
appearance and correspondence differently across model scales.
We now examine whether Gram anchoring, introduced in
DINOv3~\citep{simeoni2025dinov3}, can alleviate these effects.
By constraining pairwise patch similarities, Gram anchoring
provides a useful intervention for testing whether preserving
relational structure also preserves recoverable appearance.

We compare DINO$_{+\mathrm{iBOT}}$, standard Gram anchoring,
and DINO-Gram using ViT-B/16 and ViT-L/16.
The Gram teacher is initialized from an epoch-250 checkpoint
of DINO$_{+\mathrm{iBOT}}$ for standard Gram anchoring and
DINO for DINO-Gram.
Regularization begins at epoch 260, targeting the late-stage
deterioration observed in dense prediction. Both variants refresh
the Gram teacher from the current EMA teacher at epochs 270, 280,
and 290, so DINO-Gram differs in its initial anchor.
Implementation details are provided in
Appendix~\ref{app:gram_settings}.

\vspace{-0.1in}
\subsection{Downstream Gains Do Not Imply Uniform Locality Recovery}
\vspace{-0.1in}

\begin{figure}[t]
  \centering
  \includegraphics[width=\textwidth]
    {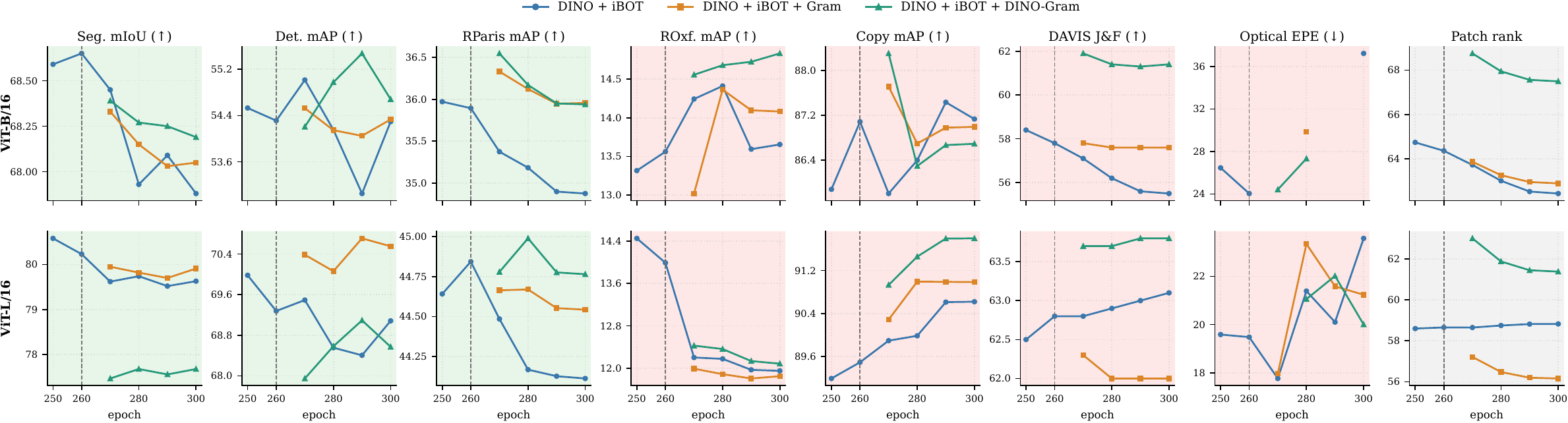}
  \caption{
  Downstream trajectories during epochs 250--300.
  Rows show ViT-B/16 and ViT-L/16; the dashed line marks
  Gram activation at epoch 260.
  Panels use independent linear vertical scales.
  Non-finite optical-flow results and EPE values above
  100 are shown as gaps. Optical flow measures training-set fitting.
  }
  \vspace{-0.2in}
  \label{fig:gram_downstream}
\end{figure}

Figure~\ref{fig:gram_downstream} shows that the downstream
effects depend on both backbone scale and teacher choice.
For ViT-B/16, the clearest improvement occurs on DAVIS:
J\&F increases from 55.5 to 57.6 with standard Gram anchoring
and to 61.4 with DINO-Gram.
Retrieval also improves, whereas segmentation and detection
change relatively little.
For ViT-L/16, standard Gram anchoring benefits segmentation
and detection but reduces DAVIS performance.
DINO-Gram instead improves DAVIS and copy detection while
sacrificing some segmentation and detection performance.

This task dependence extends the observation in
Section~\ref{sec:phenomenon}: dense semantic prediction and
fine-grained correspondence do not measure interchangeable
properties.
In particular, stabilizing dense prediction does not establish
that lost local evidence has been restored.
The two annotation-free diagnostics reveal which aspects
of locality change under each intervention.

\vspace{-0.1in}
\subsection{Teacher Choice Affects Appearance and Correspondence Differently}
\vspace{-0.1in}

\begin{figure}[t]
  \centering
  \includegraphics[width=\textwidth]
    {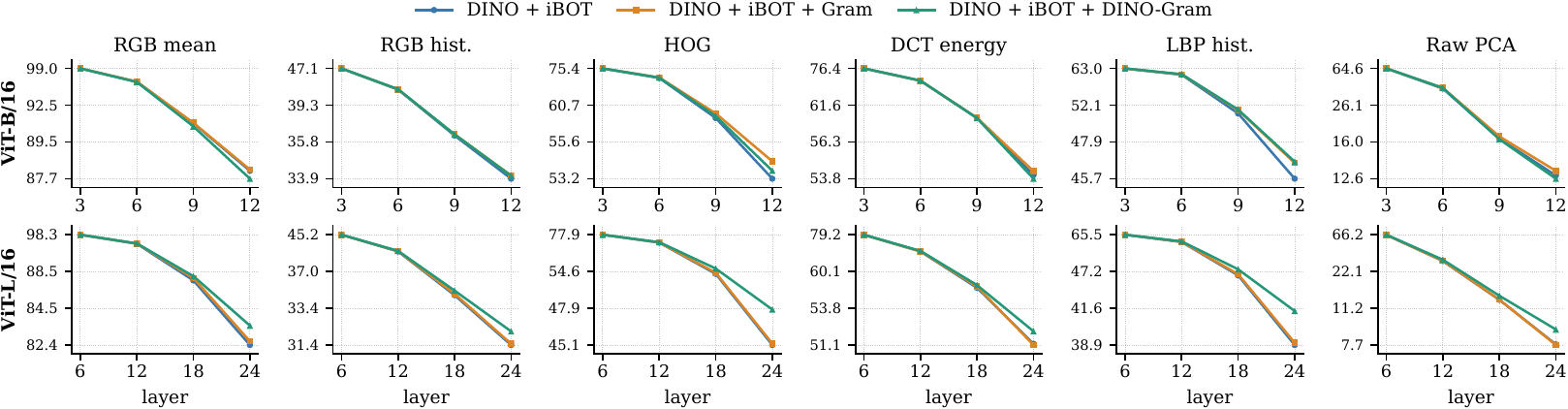}
  \caption{
  Local Appearance Decodability across layers at epoch 300.
  DINO-Gram improves deep-layer decodability more consistently
  for ViT-L/16 than for ViT-B/16.
  Values are $R^2\times100$; nonlinear asinh axes
  emphasize deep-layer differences.
  }
  \label{fig:gram_decodability}
\end{figure}

\begin{figure}[t]
  \centering
  \includegraphics[width=\textwidth]
    {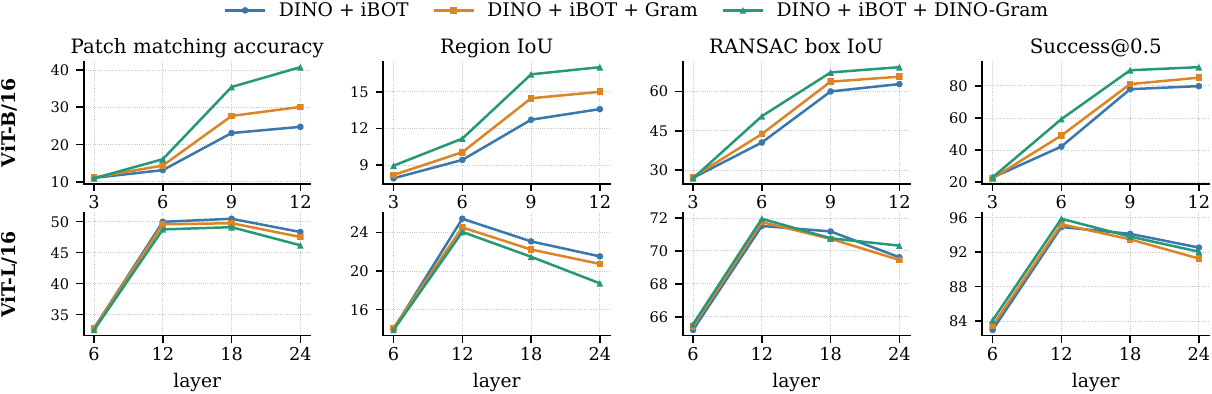}
  \caption{
  Local-to-Global Correspondence at epoch 300 with crop
  area ratio 0.1.
  Both Gram variants improve deep-layer correspondence
  for ViT-B/16, especially DINO-Gram.
  Neither consistently improves the four metrics for
  ViT-L/16. All values are percentages.
  }
  \label{fig:gram_correspondence}
\end{figure}

For ViT-B/16, Gram anchoring primarily changes correspondence.
Figure~\ref{fig:gram_decodability} shows small or mixed changes
in appearance decodability, whereas
Figure~\ref{fig:gram_correspondence} shows clear improvements
in deeper-layer correspondence.
Final-layer patch matching accuracy increases from 24.8\%
to 30.1\% with standard Gram anchoring and to 40.8\%
with DINO-Gram.
The accompanying DAVIS gains indicate that this improvement
extends beyond the diagnostic.
This is consistent with the earlier finding that iBOT's
main locality cost in smaller models concerns patch
discriminability rather than recoverable pixel and texture
evidence. Improving correspondence at this scale need not
require a broad increase in appearance decodability.

For ViT-L/16, the relationship reverses.
Standard Gram anchoring changes appearance decodability
little, while DINO-Gram improves all six final-layer
descriptors, particularly HOG and LBP.
However, neither variant consistently improves correspondence:
DINO-Gram reduces patch matching accuracy and region IoU,
despite a small increase in RANSAC box IoU.
Even with DINO-Gram, all six final-layer descriptor scores remain
below those of DINO without iBOT, indicating partial appearance recovery.
Thus, the intervention can improve the property that iBOT
weakens in larger models without improving cross-view
patch matching.

\vspace{-0.1in}
\subsection{What Does Gram Anchoring Preserve?}
\vspace{-0.1in}

A possible explanation is that Gram anchoring preserves
the teacher's organization of patch similarities rather
than local information in a general sense.
An earlier DINO$_{+\mathrm{iBOT}}$ teacher may restrain
subsequent representational drift, but its similarities
may already reflect the appearance or correspondence
limitations induced by iBOT.
An anchor can therefore stabilize a representation
without supplying the evidence that it already lacks.

Changing the teacher to DINO may provide a different
relational target.
Our earlier diagnostics suggest why this change could
have different effects across scales:
DINO retains stronger correspondence in smaller models
and stronger deep-layer appearance decodability in larger
models.
The observed benefits of DINO-Gram follow this distinction.
We therefore hypothesize that the usefulness of an anchor
depends on which property its teacher preserves, rather
than on spectral diversity alone.
Indeed, DINO-Gram increases patch effective rank at both
scales, but improves patch matching only for ViT-B/16.

Finally, within-view similarity preservation does not
directly enforce cross-view feature alignment.
Gram matrices constrain relations among patches within
an image, while correspondence additionally requires those
relations to support consistent matching across views.
Likewise, appearance-dependent similarities may retain
texture or color differences without making them invariant
to changes in viewpoint or context.
This distinction offers a possible explanation for the
ViT-L/16 result: improved appearance decodability can
coexist with weaker patch matching.
These mechanisms remain hypotheses, but the observed
dissociation shows that Gram anchoring should be assessed
separately for appearance preservation, correspondence,
and downstream utility.

\vspace{-0.1in}
\section{Conclusion}
\label{sec:conclusion}

This work shows that the locality cost of masked patch
self-distillation cannot be understood through a single measure
of local feature quality. Spatial correspondence and recoverable
appearance can respond differently to the same objective, and
scaling can improve one while weakening the other. This distinction
changes how progress toward general-purpose visual representations
should be assessed: semantic transfer and visually coherent feature
maps provide valuable evidence, but neither establishes that the
representations retain the full range of local cues needed by
downstream systems.

Our findings motivate training and evaluation practices that
explicitly account for these complementary requirements.
Future objectives could adapt their preservation targets to model
scale and depth, while teacher selection could consider both
appearance accessibility and correspondence rather than a single
downstream score. The Gram anchoring results make this direction
particularly relevant: preserving a teacher's relational structure
inherits its preferences, which may benefit different local
abilities unequally. For visual backbones serving multimodal and
spatial reasoning systems, the broader challenge is to learn useful
semantic invariances while keeping task-relevant visual distinctions
accessible. The diagnostics developed here provide a way to
measure that balance and guide its design.

\section{Acknowledgments}

This research is supported by the National Research Foundation Singapore under its AI Singapore Programme (Award Number: AISG3-RP-2022-030). Any opinions, findings, and conclusions expressed in this material are those of the author(s) and do not reflect the views of funding bodies.

\bibliographystyle{plainnat}
\bibliography{references}

\clearpage
\appendix
\part*{Appendix}
\addcontentsline{toc}{part}{Appendix}

\etocsetnexttocdepth{subsection}
{\small \localtableofcontents}
\clearpage
\section{Class-Token Joint-Embedding Objectives}
\label{sec:cls_objectives}

For a controlled comparison between DINO and DINO$_{+\mathrm{iBOT}}$,
we use the same DINO loss configuration in both variants,
following DINOv2~\cite{oquab2023dinov2}. In particular,
we construct the DINO teacher targets using Sinkhorn--Knopp
normalization, as employed in SwAV~\cite{2020_NIPS_Caron}
and DINOv2. Below, we detail this objective using the
notation introduced in Section~\ref{sec:cls_self_distillation}.

\paragraph{Student predictions.}
For a student crop $u\in\mathcal C_i$, the class-token
head produces logits $\bm s_i^{u}\in\mathbb R^K$.
The student prediction is obtained by applying
temperature-scaled softmax:
\begin{equation}
  p_{i,k}^{u}
  =
  \frac{
    \exp\big(s_{i,k}^{u}/\tau_s\big)
  }{
    \sum_{j=1}^{K}
    \exp\big(s_{i,j}^{u}/\tau_s\big)
  },
  \qquad k=1,\ldots,K,
  \label{eq:dino_student_prediction}
\end{equation}
where $\tau_s>0$ is the student temperature and $K$
is the output dimension of the class-token head.

\paragraph{Teacher targets.}
The teacher processes only global crops.
We collect their logits $\bm t_i^{v}\in\mathbb R^K$,
$v\in\mathcal G_i$, into a matrix
$\bm T\in\mathbb R^{N\times K}$, where
\begin{equation}
  N
  =
  \sum_{i=1}^{B}|\mathcal G_i|
\end{equation}
is the total number of teacher crops in the mini-batch.
Each row of $\bm T$ corresponds to one teacher crop.
We construct the target matrix as
\begin{equation}
  \bm Q
  =
  \sg\!\left[
    \operatorname{SK}
    \left(
      \exp\left(\frac{\bm T}{\tau_t}\right)
    \right)
  \right],
  \label{eq:dino_sk_targets}
\end{equation}
where the exponential is applied elementwise,
$\tau_t>0$ is the teacher temperature, and
$\sg[\cdot]$ denotes the stop-gradient operation.

The Sinkhorn--Knopp operator $\operatorname{SK}$
alternately rescales rows and columns to produce
per-crop probability distributions with balanced
aggregate assignment mass across output dimensions.
Under our scaling convention, the desired marginals are
\begin{equation}
  \bm Q\bm 1_K
  =
  \bm 1_N,
  \qquad
  \bm Q^{\top}\bm 1_N
  =
  \frac{N}{K}\bm 1_K,
  \label{eq:dino_sk_constraints}
\end{equation}
where $\bm 1_d$ is the $d$-dimensional all-ones vector.
In practice, a finite number of normalization iterations
is used, ending with row normalization. Thus, each row
sums to one, while the column-marginal constraint is
satisfied approximately. In distributed training,
the balancing statistics are aggregated across devices.
The row corresponding to teacher crop $v$ of image
$x_i$ defines its target distribution $\bm q_i^{v}$.
The temperature $\tau_t$ controls the sharpness of
these assignments.

\paragraph{Cross-view self-distillation loss.}
For each image, we match every global teacher target
to student predictions from the other crops.
The valid student--teacher pairs are
\begin{equation}
  \mathcal P_i
  =
  \big\{
    (u,v):
    u\in\mathcal C_i,\;
    v\in\mathcal G_i,\;
    u\neq v
  \big\},
  \label{eq:dino_view_pairs}
\end{equation}
where $u\neq v$ excludes pairing the same crop across
the two networks. The DINO objective is the average
cross-entropy over images and valid view pairs:
\begin{equation}
  \mathcal L_{\mathrm{base}}
  =
  \LDINO
  =
  -\frac{1}{B}
  \sum_{i=1}^{B}
  \frac{1}{|\mathcal P_i|}
  \sum_{(u,v)\in\mathcal P_i}
  \sum_{k=1}^{K}
  q_{i,k}^{v}\log p_{i,k}^{u}.
  \label{eq:cls_dino_loss}
\end{equation}
Gradients propagate only through the student predictions.
The teacher targets, including the Sinkhorn--Knopp
normalization, are computed without gradient propagation,
and the teacher parameters are updated by the exponential
moving average described in
Section~\ref{sec:cls_self_distillation}.
\section{Downstream Benchmarks}

\subsection{Capabilities Probed by the Downstream Benchmarks}
\label{app:benchmark_capabilities}

The benchmarks place different demands on semantic discrimination,
appearance information, and spatial correspondence. These categories
describe their principal evaluation targets rather than mutually
exclusive sources of information.

\paragraph{Semantic recognition and localization.}
ImageNet $k$-NN classification evaluates category discrimination using
global image features, without explicitly testing localization or
correspondence. Local cues may nevertheless contribute to recognition.
VOC2012 segmentation~\cite{everingham2015pascal} and VOC2007
detection~\cite{2010_IJCV_VOC} additionally require spatially accurate
masks or boxes. Their targets emphasize category-correct localization:
segmentation assigns the same label to visually different patches
within an object, while detection does not evaluate its internal
appearance. Consequently, improvements on these tasks do not establish
better preservation of pixel or texture information.

\paragraph{RParis and ROxford instance retrieval.}
Both benchmarks require retrieving the same landmark across images,
rather than merely recognizing its semantic category
\cite{radenovic2018revisiting}. They therefore combine global
landmark structure with distinctive local evidence. We report them
separately because their responses to iBOT differ: RParis improves
on ViT-S/L/H, whereas ROxford declines on ViT-S/B/L and remains
nearly unchanged on ViT-H. One possible explanation is greater
sensitivity to fine-grained instance discrimination on ROxford under
our protocol. This remains an interpretation of the observed results,
rather than evidence that RParis is purely semantic or ROxford
exclusively local. Both use the Hard retrieval protocol.

\paragraph{Copydays copy detection.}
Copydays tests whether transformed images can be identified as copies
of the same source~\cite{douze2009evaluation}. Category similarity is
insufficient; source-specific visual evidence must remain recognizable
under transformations. However, our protocol compares aggregated
image descriptors, so performance does not directly measure either
pixel reconstruction or patch-to-patch matching.

\paragraph{DAVIS video object tracking.}
Our DAVIS evaluation propagates an initial object mask through
similarities between frozen patch features across frames
\cite{caron2021dino,ponttuset2017davis}. It therefore emphasizes
cross-frame matching and preservation of instance identity.
Semantic coherence can assist propagation, but distinguishing
same-category instances requires more specific information.
The modest ViT-L/H gains motivate testing whether iBOT improves
correspondence at larger scales; they do not independently establish
this mechanism because tracking also depends on mask propagation
and temporal aggregation.

\paragraph{Sintel optical flow.}
Sintel evaluates dense motion estimation, including accurate
displacements and motion boundaries~\cite{butler2012naturalistic}.
Appearance cues support matching, while context helps resolve
ambiguity and occlusion. Pixel-level motion predictions do not require
reconstructing pixel values, and our learned flow head introduces an
additional readout dependency. Our flow head is evaluated on the same
videos used to fit it, so this is a training-set readout diagnostic.
The plotted comparisons retain EPE
values at most 100 and use the latest common retained checkpoint:
epochs 180, 260, 300, and 270 for ViT-S/B/L/H, respectively.
They are therefore not uniformly final-epoch comparisons.
The threshold is a reporting filter, not a definition of a valid EPE
or a guarantee that all excluded evaluations failed numerically.

\paragraph{Patch effective rank.}
Effective rank measures the spread of the singular-value spectrum of
the uncentered patch-token matrix~\cite{roy2007effective,garrido2023rankme}.
A decrease indicates greater concentration, but does not identify
which information is lost or whether the remaining directions support
better matching. We interpret it as a complementary spectral
statistic rather than a direct measure of local capability.

\subsection{Experimental Settings}
\label{app:experimental_settings}

This appendix documents the full pretraining and evaluation protocol used in
Section~\ref{sec:phenomenon}. All pretraining runs use the
DINOv2 codebase~\citep{oquab2023dinov2}.

\vspace{-0.1in}
\subsubsection{Pretraining}
\label{app:pretrain_settings}
\vspace{-0.05in}

\paragraph{Backbones.}
We compare DINO and DINO$_{+\mathrm{iBOT}}$ using ViT-S/16,
ViT-B/16, ViT-L/16, and ViT-H/16, implemented within the
DINOv2 training framework~\citep{oquab2023dinov2}.
The four backbones contain 12, 12, 24, and 32 transformer blocks,
respectively. All models use a patch size of 16, SwiGLU feed-forward
layers, four register tokens, and LayerScale initialized to
$10^{-5}$. We use uniform stochastic depth across blocks, with a
drop-path rate of $0.3$ for ViT-S/16 and ViT-B/16 and $0.4$ for
ViT-L/16 and ViT-H/16. The teacher shares the student's architecture
and is updated by an exponential moving average, with momentum
increasing from $0.994$ to $1.0$ according to a cosine schedule.

\vspace{-0.1in}
\paragraph{Training objectives.}
Both methods use the DINO class-token self-distillation
objective~\citep{caron2021dino}.
DINO$_{+\mathrm{iBOT}}$ additionally uses the masked patch
self-distillation objective of iBOT~\citep{zhou2022ibot},
with its loss weight fixed at $1.0$ throughout training.
For each backbone, the paired runs share the same architecture,
optimization settings, and data augmentation recipe; the intervention
is the inclusion of the iBOT branch and its associated masking.

To reduce direct coupling between the class-token and patch-token heads,
we follow the DINOv2 setup~\citep{oquab2023dinov2} and use separate
projection heads for the two branches. Both objectives still update
the shared backbone.
Each head consists of a three-layer MLP with hidden dimension $2048$
and bottleneck dimension $256$, followed by an output layer with
$131{,}072$ prototypes. The teacher temperature increases linearly
from $0.04$ to $0.07$ over the first $30$ epochs and remains at
$0.07$ thereafter. The final projection layers are frozen during
the first epoch. The student temperature is $0.1$ for both heads.
For iBOT, we apply the standard settings: the masking probability is $0.5$,
with masking ratios sampled within $[0.1,0.5]$.
All models additionally apply the KoLeo
regularizer~\citep{sablayrolles2018spreading} to the class-token
features with a weight of $0.1$.

\vspace{-0.1in}
\paragraph{Optimization and data augmentation.}
All models are pretrained from scratch on
ImageNet-1K~\citep{2009_ImageNet} for $300$ epochs using AdamW
with $\beta=(0.9,0.999)$.
The global batch size is $1024$ for ViT-S/16, ViT-B/16, and
ViT-L/16. Due to GPU memory constraints, both ViT-H/16 runs use
a global batch size of $512$.
The peak learning rate follows the square-root scaling rule
\[
    \eta_{\mathrm{peak}}
    = 4\times10^{-3}\sqrt{B/1024},
\]
where $B$ is the global batch size. This gives a peak learning
rate of $4\times10^{-3}$ for ViT-S/16, ViT-B/16, and ViT-L/16,
and approximately $2.83\times10^{-3}$ for ViT-H/16.
We use a $10$-epoch linear warmup followed by cosine decay to
$10^{-6}$. Weight decay increases from $0.04$ to $0.4$ according
to a cosine schedule. The gradient-clipping threshold is $3.0$,
the patch-embedding learning-rate multiplier is $0.2$, and
the layerwise learning-rate decay factor is $0.9$.

Each image produces two global crops of size $224\times224$
and eight local crops of size $96\times96$.
The crop scale ranges are $[0.32,1.0]$ for global crops and
$[0.05,0.32]$ for local crops.
All runs use four NVIDIA H200 GPUs with \texttt{bf16} mixed
precision and Fully Sharded Data Parallel (FSDP) training.
The per-GPU batch size is $256$ for ViT-S/16, ViT-B/16,
and ViT-L/16, and $128$ for ViT-H/16.

\vspace{-0.1in}
\paragraph{Compute resources.}
Table~\ref{tab:compute_resources_pretraining} summarizes the
approximate peak GPU memory usage and wall-clock training time.
Adding iBOT increases both memory consumption and training time
at every backbone scale. A substantial source of this overhead
is the patch-level prediction of distributions over $131{,}072$
prototypes, which requires large student logits and teacher
targets for the masked patches, together with the associated
training buffers. Consequently, the additional memory cost
depends on the number of processed patches as well as the
backbone size. The eight runs in Table~\ref{tab:compute_resources_pretraining}
sum to approximately 3,964 H200 GPU-hours, excluding Gram experiments,
downstream fitting, diagnostics, and exploratory runs.
In particular, the ViT-H/16 measurements use
half the batch size of the other backbones and should be
interpreted with this difference in mind.

\begin{table}[H]
  \centering
  \small
  \caption{Compute resources for 300-epoch pretraining on
    ImageNet-1K. All runs use four NVIDIA H200 GPUs with
    \texttt{bf16} mixed precision and FSDP.
    Batch size denotes the global batch size.
    Memory is the approximate peak usage per GPU, and time
    is the approximate wall-clock duration of a complete run.}
  \label{tab:compute_resources_pretraining}
  \renewcommand{\arraystretch}{1.15}
  \begin{tabular}{llrrl}
    \toprule
    \textbf{Backbone} & \textbf{Objective}
      & \textbf{Batch size} & \textbf{Memory (GB)}
      & \textbf{Time} \\
    \midrule
    ViT-S/16 & DINO
      & 1024 & 28  & 1\,d 7\,h \\
             & DINO$_{+\mathrm{iBOT}}$
      & 1024 & 100 & 2\,d 5\,h \\
    \midrule
    ViT-B/16 & DINO
      & 1024 & 47  & 3\,d 1\,h \\
             & DINO$_{+\mathrm{iBOT}}$
      & 1024 & 125 & 3\,d 8\,h \\
    \midrule
    ViT-L/16 & DINO
      & 1024 & 100 & 5\,d 15\,h \\
             & DINO$_{+\mathrm{iBOT}}$
      & 1024 & 129 & 6\,d 5\,h \\
    \midrule
    ViT-H/16 & DINO
      & 512 & 86  & 9\,d 8\,h \\
             & DINO$_{+\mathrm{iBOT}}$
      & 512 & 134 & 10\,d 6\,h \\
    \bottomrule
  \end{tabular}
  \vspace{-0.1in}
\end{table}

\vspace{-0.1in}
\subsubsection{Downstream Evaluation}
\label{app:eval_settings}
\vspace{-0.1in}
We evaluate the frozen teacher checkpoints on eight
downstream benchmarks, counting RParis and ROxford separately.
Below we describe the benchmark and the protocol used
for each task.
\vspace{-0.1in}
\paragraph{ImageNet $k$-NN classification.}
We compute frozen class-token features on the ImageNet-1K training set and
evaluate weighted $k$-NN classification on the validation set with cosine
similarity. We use $k=15$ and a softmax voting
temperature of $0.07$, and report top-1 accuracy following the
DINO protocol~\citep{caron2021dino}. The eval transform is the standard
DINOv2 ImageNet evaluation transform: resize the shorter edge to $256$,
center crop $224\times224$, and normalize with ImageNet mean/std.
\vspace{-0.1in}
\paragraph{RParis and ROxford image retrieval.}
We evaluate on the revisited Paris6k and Oxford5k
benchmarks~\citep{radenovic2018revisiting}.
Database and query images are resized to $224\times224$, and we extract a
single image-level descriptor by concatenating the class token with a
generalized-mean-pooled patch-token vector with exponent $4$. We report the
mean average precision (mAP) under the Hard evaluation
protocol of~\citet{radenovic2018revisiting}. No fine-tuning or whitening is
performed.
\vspace{-0.1in}
\paragraph{VOC2012 semantic segmentation.}
We use a linear segmentation protocol implemented in MMSegmentation. The frozen ViT backbone produces patch tokens at
$512\times512$ input resolution; we concatenate features from $4$ uniformly
spaced block-chunk outputs after layer normalization and feed them to
an FCN head with no intermediate convolutions. The concatenated feature
dimensions are $1536$, $3072$, $4096$, and $5120$ for ViT-S/B/L/H,
respectively. Training uses Pascal VOC2012
\texttt{train} for $10{,}000$ iterations with AdamW, learning rate
$5\!\times\!10^{-4}$, a total batch size 16, weight decay $0$, $200$-iteration linear warmup, and a
power-1 polynomial schedule. Random scaling in $[0.5,2.0]$, $512\times512$
random crops, random horizontal flips, and photometric distortion are applied
during training. Evaluation uses sliding-window inference with $512\times512$
windows and stride $341$ on Pascal VOC2012 \texttt{val}, and we report mIoU.
\vspace{-0.1in}
\paragraph{VOC2007 object detection.}
We adopt a Cascade R-CNN\cite{cai2018cascade} detector with a $4$-conv-$1$-fc GIoU head and FPN
neck on top of the frozen ViT, implemented in MMDetection. The
backbone is fully frozen; FPN inputs are $4$
uniformly spaced block-chunk outputs after layernorm. Training uses VOC2007
\texttt{train} with $1000\times600$ short-side resize, random horizontal
flips, AdamW with learning rate $1\!\times\!10^{-4}$, a total batch size 16, weight decay $0.05$,
$500$-iteration linear warmup from a $10^{-3}$ factor, and a multi-step
decay at epochs $8$ and $11$ for $12$ training epochs. Evaluation uses the
VOC2007 \texttt{val} with the standard 11-point interpolation
mAP~\citep{2010_IJCV_VOC}.
\vspace{-0.1in}
\paragraph{Copydays copy detection.}
We follow the DINO copy-detection protocol on the Copydays
dataset~\citep{douze2009evaluation}, with $10{,}000$
COCO~\citep{2014_ECCV_COCO} distractor images and $20{,}000$ COCO whitening
images. Images are resized to $224\times224$, and we form a single
descriptor by concatenating the class token with the GeM-pooled patch tokens
(exponent $4$). We center features, fit a PCA whitening with whitening
exponent $0.5$ on the COCO whitening set, and $\ell_2$-normalize. Retrieval
uses cosine similarity, and we report mAP on \texttt{strong} subset.
\vspace{-0.1in}
\paragraph{DAVIS video object tracking.}
We follow the DINO label-propagation tracking protocol on
DAVIS-2017~\citep{ponttuset2017davis}. For each video, the first-frame mask is
propagated to subsequent frames via cosine similarity between frozen patch
tokens of the target frame and the past $7$ frames (plus the first frame).
Frames are resized so that the shorter side is $480$ and the longer side is
rounded to a multiple of $64$. We restrict matching to a spatial neighborhood
of radius $12$ patches around the query, aggregate from the top-$5$ neighbors,
and use a softmax temperature of $0.1$. We report the standard region--boundary
score $\mathcal J\&\mathcal F$ on the DAVIS-2017
\texttt{val} split using the DAVIS official evaluation
scripts~\citep{ponttuset2017davis}.
\vspace{-0.1in}
\paragraph{Sintel optical flow.}
We use the PWCNet optical-flow head from MMFlow on top of the frozen ViT
backbone with FPN. The
encoder is fully frozen; FPN-projected feature
maps with channels $\{32,64,96,128\}$ at levels $\{2,3,4,5\}$ feed the
PWCNet~\cite{sun2018pwc} decoder. Training runs for $12$ epochs on the Sintel
\texttt{clean+final} training pass with $384\times768$ random crops, random
affine transforms, color jitter, gamma jitter, random horizontal/vertical
flips, AdamW with learning rate $1\!\times\!10^{-4}$, a total batch size 16, weight decay $0.05$,
$500$-iteration linear warmup, and a multi-step decay at epochs $8$ and
$11$. We report the average end-point error (EPE) on the
\texttt{clean} pass of the same Sintel training videos, using the
evaluation preprocessing pipeline. No disjoint validation split is
used. These results measure the fitting behavior of a learned flow
head on frozen features, not held-out optical-flow generalization.
\vspace{-0.1in}
\paragraph{Patch Effective Rank.}
To quantify the diversity of patch-level representations, we compute the
normalized effective rank of the patch token matrix for each image. Given an
image $I_b$, let its patch tokens be denoted as
$
\mathbf{X}_b = [\mathbf{x}_{b,1}, \mathbf{x}_{b,2}, \ldots, \mathbf{x}_{b,P}]^\top
\in \mathbb{R}^{P \times D},
$
where $P$ is the number of patches and $D$ is the feature dimension of each
patch token. We perform singular value decomposition on $\mathbf{X}_b$ and
obtain its singular values
$\{\sigma_{b,i}\}_{i=1}^{q}$, where $q=\min(P,D)$.
The singular values are normalized as
$
p_{b,i} =
\frac{\sigma_{b,i}}{\sum_{j=1}^{q}\sigma_{b,j}},
$
which can be viewed as a probability distribution over spectral components.
The entropy of this distribution is then computed by
$
H_b = - \sum_{i=1}^{q} p_{b,i} \log p_{b,i},
$
with $0\log0=0$ and a small numerical floor in the implementation.
Following the definition of effective rank, the patch effective rank is given by
$
\operatorname{ERank}(\mathbf{X}_b) = \exp(H_b).
$
We report effective rank relative to the number of patches:
\[
\operatorname{PER}(I_b)
=
\frac{\exp\left(-\sum_{i=1}^{q} p_{b,i}\log p_{b,i}\right)}{P}.
\]

A larger value indicates a more evenly distributed singular-value
spectrum; a smaller value indicates greater spectral concentration.
Because $\mathbf X_b$ is not centered and the probabilities use singular
values rather than their squares, this statistic is not a distribution
of covariance variance. It does not by itself establish information
loss, representational collapse, or poorer correspondence.
The theoretical upper bound of the normalized score is $q/P$, which
equals one when $D\ge P$ as in our evaluated backbones.

In our experiments, the patch effective rank is computed on $10{,}000$ randomly sampled ImageNet
training images using subset seed 0. We report the mean normalized
patch effective rank across images, multiplied by 100 in plots and tables.

\subsubsection{Experimental Settings for Gram Anchoring}
\label{app:gram_settings}

\paragraph{Models and comparison.}
We evaluate Gram anchoring using ViT-B/16 and ViT-L/16
under the 300-epoch ImageNet-1K pretraining setup described
in Appendix~\ref{app:pretrain_settings}.
The reference configuration is DINO$_{+\mathrm{iBOT}}$.
The two additional configurations introduce a Gram loss
whose teacher is initialized from either the
DINO$_{+\mathrm{iBOT}}$ or DINO epoch-250 checkpoint
of the same backbone architecture.
We refer to these configurations as standard Gram
anchoring and DINO-Gram, respectively.
The DINO and iBOT objective weights remain 1.0.

\paragraph{Gram objective.}
Let $Z_s,Z_g\in\mathbb{R}^{N\times d}$ denote student
and Gram-teacher patch features for a global crop,
and let $\widehat Z_s,\widehat Z_g$ denote their
row-wise $\ell_2$-normalized versions.
The student features come from the masked global-crop forward pass,
whereas the Gram teacher processes the corresponding unmasked crop.
All patch positions, including masked student positions, enter the
Gram matrices. The objective matches within-crop pairwise similarities:
\begin{equation}
\mathcal L_{\mathrm{Gram}}
=
\frac{1}{N^2}
\left\|
\widehat Z_s\widehat Z_s^\top
-
\widehat Z_g\widehat Z_g^\top
\right\|_F^2,
\end{equation}
averaged over global crops in the batch.
The loss uses backbone patch features rather than
prototype-head outputs. We do not subtract the spatial
feature mean or clamp negative similarities.
Gradients propagate through the student features only.

\paragraph{Activation and teacher schedule.}
Gram regularization is inactive for the first 260 epochs.
Its coefficient then increases from 0 to 2 over one epoch
using a cosine transition and remains 2 thereafter.
The initial Gram-teacher checkpoint is taken at epoch 250,
before the late-stage deterioration motivating the intervention.
For both standard Gram anchoring and DINO-Gram, we perform
three subsequent teacher updates at epochs 270, 280,
and 290. Each update copies the current EMA teacher
into the Gram teacher. The Gram teacher is therefore
fixed between updates, rather than fixed throughout
the entire intervention. Thus, DINO-Gram uses DINO for initialization;
its later anchors are the current run's EMA teacher, not a permanently
fixed DINO checkpoint.

\paragraph{Optimization.}
Both backbones use a global batch size of 1024,
distributed across four GPUs.
The drop-path rates are 0.3 for ViT-B/16 and 0.4
for ViT-L/16.
The AdamW optimizer, learning-rate and weight-decay
schedules, crop sizes, and other pretraining settings
follow Appendix~\ref{app:pretrain_settings}.
Gram regularization is added to the existing objective:
\begin{equation}
\mathcal L
=
\mathcal L_{\mathrm{DINO}}
+
\mathcal L_{\mathrm{iBOT}}
+
0.1\,\mathcal L_{\mathrm{KoLeo}}
+
\lambda_{\mathrm{Gram}}(t)\mathcal L_{\mathrm{Gram}}.
\end{equation}

\paragraph{Evaluation and plotting.}
Downstream trajectories display the recorded checkpoints
between epochs 250 and 300, with independent linear
vertical scales for each panel.
Unavailable evaluations are not imputed.
For optical flow, non-finite EPE values and values
exceeding 100 are omitted from the displayed curves,
with gaps retained between valid observations.
This filtering does not establish successful optical-flow
evaluation for checkpoints whose results are omitted.

Both annotation-free layer sweeps use epoch 300,
without averaging across training checkpoints.
We evaluate blocks 3, 6, 9, and 12 for ViT-B/16 and
blocks 6, 12, 18, and 24 for ViT-L/16.
Appearance decodability reports the dimension-averaged
$R^2$ separately for each descriptor group.
Correspondence uses crop area ratio 0.1 and reports
patch matching accuracy, region IoU, RANSAC box IoU,
and success at IoU 0.5.
The plots compare recorded point estimates and do not
imply statistical significance.

\section{Additional Similarity Heatmap Visualization}

\begin{figure}[H]
  \centering
  \includegraphics[width=12cm]{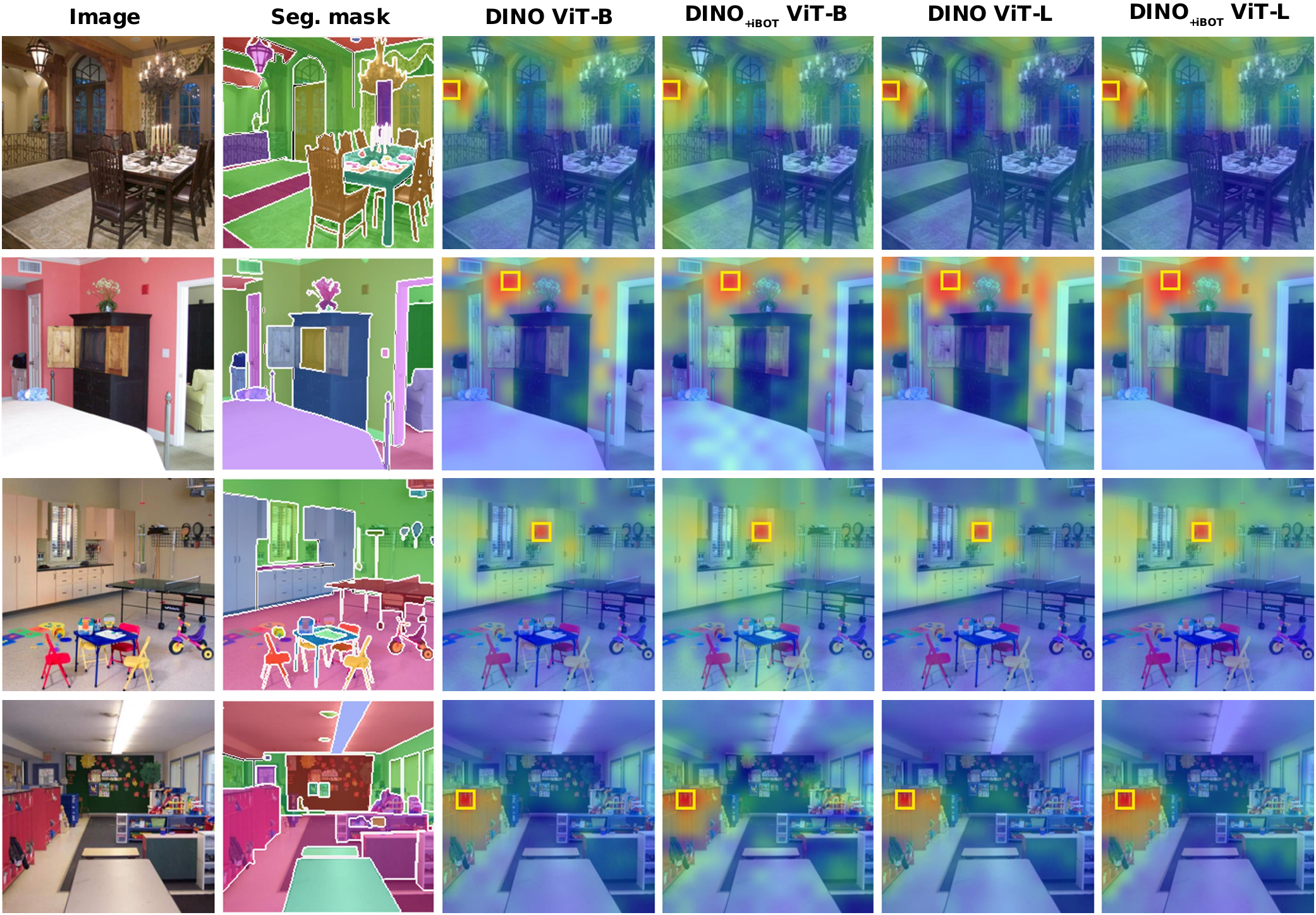}
  \caption{Additional anchor-patch similarity heatmaps for DINO and DINO$_{+\mathrm{iBOT}}$ using ViT-B/16 and ViT-L/16.}
  \label{fig:heatmap_extra}
\end{figure}

\section{Numerical Results and Trajectories of Downstream Evaluations}
\label{app:fig2_values}

Table~\ref{tab:fig2_vit_base_values} reports the numerical values
underlying Figure~\ref{fig:paired_delta} for the four
DINO/DINO$_{+\mathrm{iBOT}}$ pairs.

\begin{table}[p]
\centering
\small
\setlength{\tabcolsep}{7pt}
\renewcommand{\arraystretch}{1.2}
\begin{tabular}{l|llrrr}
\toprule
Arch. & Metric & Better & DINO
& DINO$_{+\mathrm{iBOT}}$ & $\Delta$ \\
\midrule

\multirow{9}{*}{ViT-S/16}
& \cellcolor{semanticbg} kNN top-1
& \cellcolor{semanticbg} $\uparrow$
& \cellcolor{semanticbg} 70.88
& \cellcolor{semanticbg} 71.27
& \cellcolor{semanticbg} \textcolor{improvegreen}{+0.39} \\
& \cellcolor{semanticbg} VOC2012 Seg. mIoU
& \cellcolor{semanticbg} $\uparrow$
& \cellcolor{semanticbg} 53.13
& \cellcolor{semanticbg} 56.02
& \cellcolor{semanticbg} \textcolor{improvegreen}{+2.89} \\
& \cellcolor{semanticbg} VOC2007 Det. mAP
& \cellcolor{semanticbg} $\uparrow$
& \cellcolor{semanticbg} 42.88
& \cellcolor{semanticbg} 43.96
& \cellcolor{semanticbg} \textcolor{improvegreen}{+1.08} \\
& \cellcolor{semanticbg} RParis Hard mAP
& \cellcolor{semanticbg} $\uparrow$
& \cellcolor{semanticbg} 29.75
& \cellcolor{semanticbg} 30.78
& \cellcolor{semanticbg} \textcolor{improvegreen}{+1.03} \\
& \cellcolor{localitybg} ROxford Hard mAP
& \cellcolor{localitybg} $\uparrow$
& \cellcolor{localitybg} 13.34
& \cellcolor{localitybg} 12.03
& \cellcolor{localitybg} \textcolor{degradered}{-1.31} \\
& \cellcolor{localitybg} Copy strong mAP
& \cellcolor{localitybg} $\uparrow$
& \cellcolor{localitybg} 84.14
& \cellcolor{localitybg} 83.69
& \cellcolor{localitybg} \textcolor{degradered}{-0.45} \\
& \cellcolor{localitybg} DAVIS J\&F
& \cellcolor{localitybg} $\uparrow$
& \cellcolor{localitybg} 60.10
& \cellcolor{localitybg} 59.30
& \cellcolor{localitybg} \textcolor{degradered}{-0.80} \\
& \cellcolor{localitybg} Sintel clean EPE
& \cellcolor{localitybg} $\downarrow$
& \cellcolor{localitybg} 18.19
& \cellcolor{localitybg} 18.26
& \cellcolor{localitybg} \textcolor{degradered}{+0.06} \\
& \cellcolor{localitybg} Normalized patch rank
& \cellcolor{localitybg} --
& \cellcolor{localitybg} 69.10
& \cellcolor{localitybg} 62.85
& \cellcolor{localitybg} $-6.25$ \\
\midrule

\multirow{9}{*}{ViT-B/16}
& \cellcolor{semanticbg} kNN top-1
& \cellcolor{semanticbg} $\uparrow$
& \cellcolor{semanticbg} 76.36
& \cellcolor{semanticbg} 76.97
& \cellcolor{semanticbg} \textcolor{improvegreen}{+0.60} \\
& \cellcolor{semanticbg} VOC2012 Seg. mIoU
& \cellcolor{semanticbg} $\uparrow$
& \cellcolor{semanticbg} 65.61
& \cellcolor{semanticbg} 67.88
& \cellcolor{semanticbg} \textcolor{improvegreen}{+2.27} \\
& \cellcolor{semanticbg} VOC2007 Det. mAP
& \cellcolor{semanticbg} $\uparrow$
& \cellcolor{semanticbg} 52.64
& \cellcolor{semanticbg} 54.29
& \cellcolor{semanticbg} \textcolor{improvegreen}{+1.66} \\
& \cellcolor{semanticbg} RParis Hard mAP
& \cellcolor{semanticbg} $\uparrow$
& \cellcolor{semanticbg} 37.24
& \cellcolor{semanticbg} 34.87
& \cellcolor{semanticbg} \textcolor{degradered}{-2.36} \\
& \cellcolor{localitybg} ROxford Hard mAP
& \cellcolor{localitybg} $\uparrow$
& \cellcolor{localitybg} 14.60
& \cellcolor{localitybg} 13.65
& \cellcolor{localitybg} \textcolor{degradered}{-0.95} \\
& \cellcolor{localitybg} Copy strong mAP
& \cellcolor{localitybg} $\uparrow$
& \cellcolor{localitybg} 87.61
& \cellcolor{localitybg} 87.13
& \cellcolor{localitybg} \textcolor{degradered}{-0.48} \\
& \cellcolor{localitybg} DAVIS J\&F
& \cellcolor{localitybg} $\uparrow$
& \cellcolor{localitybg} 61.80
& \cellcolor{localitybg} 55.50
& \cellcolor{localitybg} \textcolor{degradered}{-6.30} \\
& \cellcolor{localitybg} Sintel clean EPE
& \cellcolor{localitybg} $\downarrow$
& \cellcolor{localitybg} 24.44
& \cellcolor{localitybg} 24.04
& \cellcolor{localitybg} \textcolor{improvegreen}{-0.39} \\
& \cellcolor{localitybg} Normalized patch rank
& \cellcolor{localitybg} --
& \cellcolor{localitybg} 70.57
& \cellcolor{localitybg} 62.44
& \cellcolor{localitybg} $-8.13$ \\
\midrule

\multirow{9}{*}{ViT-L/16}
& \cellcolor{semanticbg} kNN top-1
& \cellcolor{semanticbg} $\uparrow$
& \cellcolor{semanticbg} 79.36
& \cellcolor{semanticbg} 80.44
& \cellcolor{semanticbg} \textcolor{improvegreen}{+1.08} \\
& \cellcolor{semanticbg} VOC2012 Seg. mIoU
& \cellcolor{semanticbg} $\uparrow$
& \cellcolor{semanticbg} 71.93
& \cellcolor{semanticbg} 79.63
& \cellcolor{semanticbg} \textcolor{improvegreen}{+7.70} \\
& \cellcolor{semanticbg} VOC2007 Det. mAP
& \cellcolor{semanticbg} $\uparrow$
& \cellcolor{semanticbg} 59.71
& \cellcolor{semanticbg} 69.08
& \cellcolor{semanticbg} \textcolor{improvegreen}{+9.38} \\
& \cellcolor{semanticbg} RParis Hard mAP
& \cellcolor{semanticbg} $\uparrow$
& \cellcolor{semanticbg} 40.81
& \cellcolor{semanticbg} 44.11
& \cellcolor{semanticbg} \textcolor{improvegreen}{+3.30} \\
& \cellcolor{localitybg} ROxford Hard mAP
& \cellcolor{localitybg} $\uparrow$
& \cellcolor{localitybg} 12.86
& \cellcolor{localitybg} 11.95
& \cellcolor{localitybg} \textcolor{degradered}{-0.91} \\
& \cellcolor{localitybg} Copy strong mAP
& \cellcolor{localitybg} $\uparrow$
& \cellcolor{localitybg} 89.74
& \cellcolor{localitybg} 90.62
& \cellcolor{localitybg} \textcolor{improvegreen}{+0.87} \\
& \cellcolor{localitybg} DAVIS J\&F
& \cellcolor{localitybg} $\uparrow$
& \cellcolor{localitybg} 62.90
& \cellcolor{localitybg} 63.10
& \cellcolor{localitybg} \textcolor{improvegreen}{+0.20} \\
& \cellcolor{localitybg} Sintel clean EPE
& \cellcolor{localitybg} $\downarrow$
& \cellcolor{localitybg} 22.84
& \cellcolor{localitybg} 23.57
& \cellcolor{localitybg} \textcolor{degradered}{+0.73} \\
& \cellcolor{localitybg} Normalized patch rank
& \cellcolor{localitybg} --
& \cellcolor{localitybg} 66.42
& \cellcolor{localitybg} 58.83
& \cellcolor{localitybg} $-7.59$ \\
\midrule

\multirow{9}{*}{ViT-H/16}
& \cellcolor{semanticbg} kNN top-1
& \cellcolor{semanticbg} $\uparrow$
& \cellcolor{semanticbg} 74.45
& \cellcolor{semanticbg} 78.11
& \cellcolor{semanticbg} \textcolor{improvegreen}{+3.65} \\
& \cellcolor{semanticbg} VOC2012 Seg. mIoU
& \cellcolor{semanticbg} $\uparrow$
& \cellcolor{semanticbg} 72.88
& \cellcolor{semanticbg} 81.95
& \cellcolor{semanticbg} \textcolor{improvegreen}{+9.07} \\
& \cellcolor{semanticbg} VOC2007 Det. mAP
& \cellcolor{semanticbg} $\uparrow$
& \cellcolor{semanticbg} 58.89
& \cellcolor{semanticbg} 68.94
& \cellcolor{semanticbg} \textcolor{improvegreen}{+10.05} \\
& \cellcolor{semanticbg} RParis Hard mAP
& \cellcolor{semanticbg} $\uparrow$
& \cellcolor{semanticbg} 35.10
& \cellcolor{semanticbg} 38.96
& \cellcolor{semanticbg} \textcolor{improvegreen}{+3.87} \\
& \cellcolor{localitybg} ROxford Hard mAP
& \cellcolor{localitybg} $\uparrow$
& \cellcolor{localitybg} 9.05
& \cellcolor{localitybg} 9.07
& \cellcolor{localitybg} \textcolor{improvegreen}{+0.02} \\
& \cellcolor{localitybg} Copy strong mAP
& \cellcolor{localitybg} $\uparrow$
& \cellcolor{localitybg} 88.74
& \cellcolor{localitybg} 88.26
& \cellcolor{localitybg} \textcolor{degradered}{-0.47} \\
& \cellcolor{localitybg} DAVIS J\&F
& \cellcolor{localitybg} $\uparrow$
& \cellcolor{localitybg} 62.20
& \cellcolor{localitybg} 62.60
& \cellcolor{localitybg} \textcolor{improvegreen}{+0.40} \\
& \cellcolor{localitybg} Sintel clean EPE
& \cellcolor{localitybg} $\downarrow$
& \cellcolor{localitybg} 24.73
& \cellcolor{localitybg} 26.66
& \cellcolor{localitybg} \textcolor{degradered}{+1.92} \\
& \cellcolor{localitybg} Normalized patch rank
& \cellcolor{localitybg} --
& \cellcolor{localitybg} 63.88
& \cellcolor{localitybg} 54.60
& \cellcolor{localitybg} $-9.29$ \\

\bottomrule
\end{tabular}
\vspace{0.1in}
\caption{
Numerical results for the four backbone scales in
Figure~\ref{fig:paired_delta}.
All rows use epoch 300 except Sintel EPE, which uses the latest common
retained checkpoint: epochs 180/260/300/270 for ViT-S/B/L/H.
Sintel scores are evaluated on training videos rather than a held-out split.
Performance metrics other than EPE are percentages.
Normalized patch rank is the mean patch effective rank
divided by the number of patches and multiplied by 100;
it describes spectral dimensionality rather than task accuracy.
$\Delta$ denotes DINO$_{+\mathrm{iBOT}}$ minus DINO,
computed before rounding.
Green and red indicate improved and degraded task performance,
respectively; patch-rank changes are shown without a
performance interpretation.
}
\label{tab:fig2_vit_base_values}
\end{table}

\begin{figure}[H]
  \centering
  \makebox[\textwidth][c]{%
    \includegraphics[width=14cm]{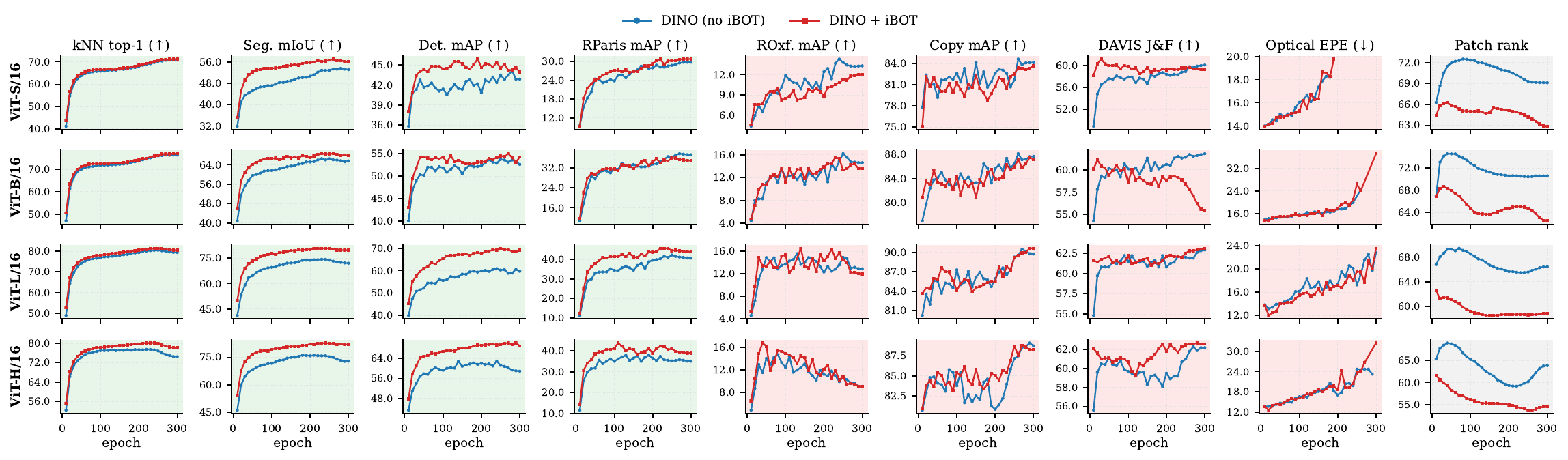}}
  \caption{
  Training trajectories of DINO and DINO$_{+\mathrm{iBOT}}$
  across four backbone scales.
  The effect of iBOT varies with the task, backbone scale,
  and training stage.
  Improvements in classification and dense semantic prediction
  coexist with mixed effects on retrieval and correspondence.
  Non-finite optical-flow results and EPE values exceeding 100
  are excluded.
  }
  \label{fig:trajectories}
  \vspace{-0.2in}
\end{figure}

At epoch 300, adding iBOT improves ImageNet kNN classification,
VOC2012 segmentation, and VOC2007 detection at all four scales.
RParis retrieval improves for ViT-S/16, ViT-L/16, and
ViT-H/16, but decreases for ViT-B/16.
The results on tasks requiring local discrimination are more mixed.
ROxford retrieval decreases for ViT-S/16, ViT-B/16, and
ViT-L/16, while remaining nearly unchanged for ViT-H/16.
Copy detection decreases at three scales, with ViT-L/16
as an exception. DAVIS exhibits a clear change with model scale:
iBOT reduces J\&F for ViT-S/16 and ViT-B/16, but yields
modest improvements for ViT-L/16 and ViT-H/16.
At the selected optical-flow checkpoints, iBOT increases
EPE for ViT-S/16, ViT-L/16, and ViT-H/16, but slightly
reduces it for ViT-B/16.

Figure~\ref{fig:trajectories} further shows that these
differences vary over training rather than following a
uniformly widening semantic--locality trade-off.
For example, the DAVIS advantage of iBOT on ViT-S/16
at earlier checkpoints reverses by epoch 300, whereas
the ViT-B/16 gap becomes more negative late in training.
Retrieval and copy-detection differences can also change sign.
Meanwhile, normalized patch rank is lower with iBOT at
epoch 300 for every backbone, including the larger models
whose DAVIS performance improves.
Thus, reduced spectral dimensionality alone does not imply
weaker correspondence. These results motivate the separate
diagnostics in Section~\ref{sec:mechanism}:
Local Appearance Decodability measures recoverable pixel
and texture evidence, whereas Local-to-Global Correspondence
measures whether patches remain spatially distinguishable
and matchable across views.

\section{Additional DAVIS Video Object Tracking Visualization}
\label{app:additional_davis_visual}

\begin{figure}[H]
  \centering
  \includegraphics[width=12cm]{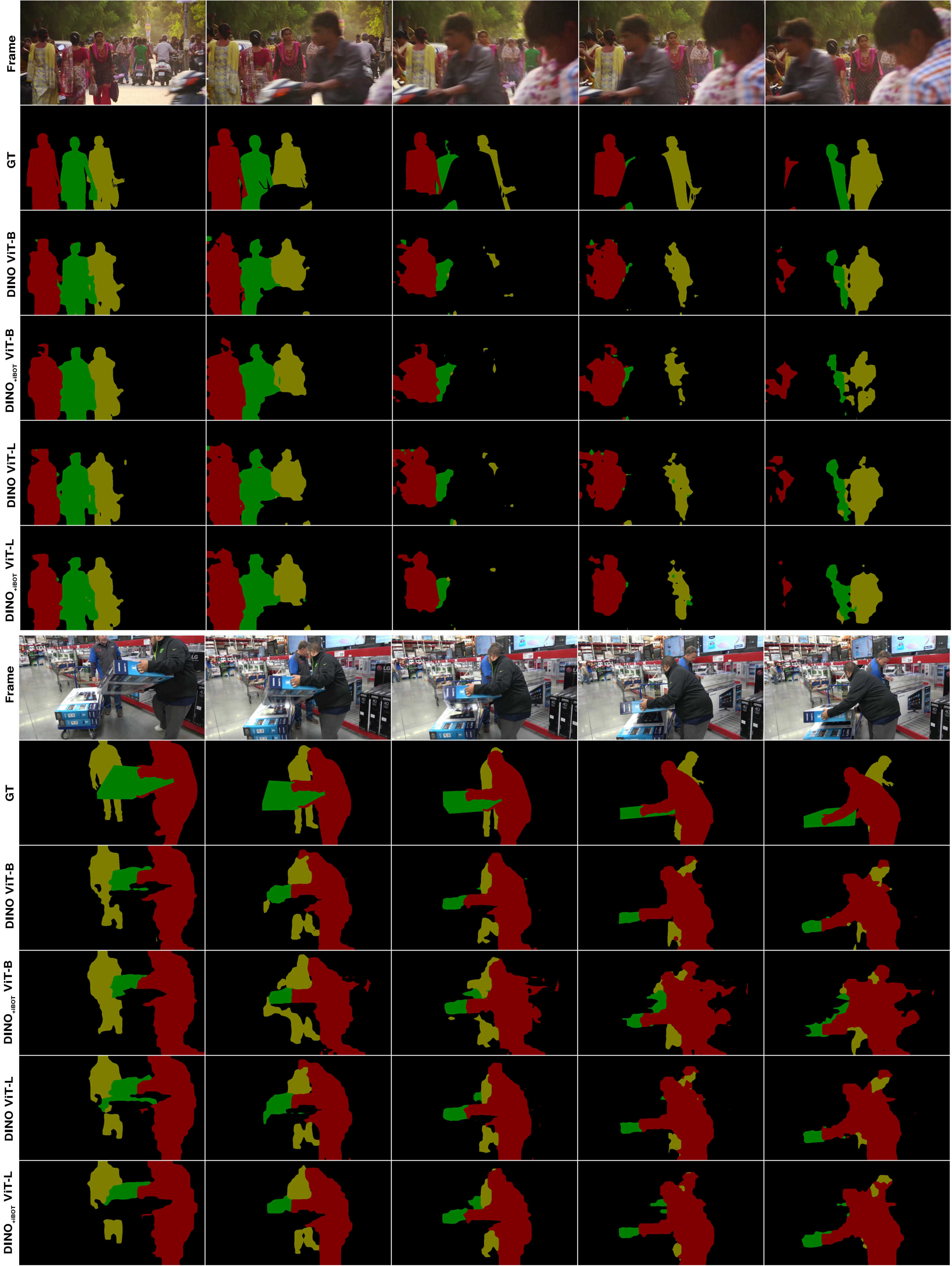}
  \caption{Additional DAVIS video object tracking visualization for DINO and DINO$_{+\mathrm{iBOT}}$ using ViT-B/16 and ViT-L/16.}
  \label{fig:davis_extra}
\end{figure}

\section{Implementation and Visualization of Local Appearance Decodability}
\label{app:descriptor_decodability}

Local Appearance Decodability asks a simple question:
\emph{how accurately can a linear decoder recover the appearance
of an image patch from its frozen patch token?}
We describe each raw patch using six groups of pixel-derived
attributes, covering color, edge orientation, texture, spatial
frequency, and raw-pixel variation. We then train a linear decoder
to predict these attributes and evaluate it on held-out images.
The backbone remains frozen throughout this procedure, and no
semantic annotations or model-generated targets are used.

\subsection{Image Preparation and Probe Data}

\paragraph{Aligning patch tokens with raw pixels.}
Each image undergoes a deterministic evaluation transform consisting
of bicubic resizing of the shorter edge to 256, a $224\times224$
center crop, and conversion to an RGB tensor
$w_{\mathrm{raw}}\in[0,1]^{3\times H\times W}$.
We use this tensor to construct the appearance targets.
A copy of the same tensor is normalized using the ImageNet mean
and standard deviation and passed to the backbone.
Thus, the backbone input and the target image have exactly the
same spatial crop.

For a backbone with patch size $p$, we divide the raw image into
non-overlapping $p\times p$ patches.
The raw patch at location $r$ is denoted by
$\tilde w_r\in[0,1]^{3\times p\times p}$, and its corresponding
token from layer $\ell$ is
$z_{\theta}^{(\ell)}(w,r)\in\mathbb{R}^{d}$.
We check that the number of raw patches matches the number of
patch tokens.

\paragraph{Separating training and validation images.}
We randomly select 20,000 ImageNet-1K training images using subset seed 0 and
divide them into 10,000 probe-training images and 10,000 validation
images. We use all $14\times14=196$ patches per image, yielding
1,960,000 token--target pairs per split. Splitting at the image level
ensures that patches
from the same image cannot appear in both sets.
All fitted preprocessing statistics, including the raw-pixel PCA
basis and the decoder's normalization statistics, are estimated
using only the probe-training data.

The MAE and SimMIM curves use separate pretrained ViT-B/16 reference
checkpoints with the same probe protocol. They are not controlled
300-epoch DINO ablations: their pretraining objectives and training
budgets differ. Their role is to contextualize appearance decodability,
not to isolate the causal effect of a pretraining objective.

\subsection{Local Appearance Targets}

At each patch location, we concatenate six descriptor groups:
\begin{equation}
  y(w,r)=\phi(w_{\mathrm{raw}},r)
  =
  \big[
    \phi_{\mathrm{rgb\_mean}},
    \phi_{\mathrm{rgb\_hist}},
    \phi_{\mathrm{hog}},
    \phi_{\mathrm{dct}},
    \phi_{\mathrm{lbp}},
    \phi_{\mathrm{raw\_pca}}
  \big](w_{\mathrm{raw}},r).
  \label{eq:appendix_appearance_concat}
\end{equation}
Their dimensions are 3, 24, 9, 6, 16, and 64, respectively,
giving a 122-dimensional target.
Although the groups are concatenated for fitting, we evaluate
and report each group separately. RGB statistics, DCT, and raw PCA
use only the aligned raw patch. HOG and LBP are computed on the full
image and then pooled within each patch, so their boundary responses
can depend on a one-pixel neighborhood outside the patch.

\paragraph{RGB mean: average color and brightness.}
We average the pixel values within each patch independently
for the red, green, and blue channels:
\begin{equation}
  \phi_{\mathrm{rgb\_mean}}^{c}(\tilde w_r)
  =
  \frac{1}{p^2}\sum_{a,b}\tilde w_r^c(a,b),
  \qquad c\in\{R,G,B\}.
  \label{eq:appendix_rgb_mean}
\end{equation}
These three values describe the patch's average color, but do
not retain the arrangement of pixels within it.

\paragraph{RGB histogram: distribution of color intensities.}
For each color channel, we divide $[0,1]$ into eight equal-width
bins and count the fraction of pixels falling into each bin.
A pixel value $v$ is assigned to bin
$\min(\lfloor 8v\rfloor,7)$.
Concatenating the three normalized histograms produces 24 values.
Unlike the RGB mean, this descriptor distinguishes patches with
similar average colors but different intensity distributions.

\paragraph{Grayscale descriptors.}
For HOG, DCT, and LBP, we convert the raw image to grayscale using
\begin{equation}
  I
  =
  0.2989\,w_{\mathrm{raw}}^R
  +0.5870\,w_{\mathrm{raw}}^G
  +0.1140\,w_{\mathrm{raw}}^B.
  \label{eq:appendix_grayscale}
\end{equation}

\paragraph{HOG: distribution of edge orientations.}
Inspired by Histograms of Oriented Gradients
(HOG)~\cite{dalal2005hog}, we summarize the edge orientations
within each patch.
We compute horizontal and vertical gradients using
$3\times3$ Sobel filters scaled by $1/8$ on the full grayscale image,
with zero padding at the image boundary.
Each pixel contributes to a nine-bin histogram of unsigned
orientations in $[0,\pi)$, weighted by its gradient magnitude.

Specifically, the magnitude is
$m=\sqrt{g_x^2+g_y^2+10^{-12}}$, and the orientation is
$\theta=\operatorname{atan2}(g_y,g_x)\bmod\pi$.
The bin centers are $\nu_k=k\pi/9$.
Votes are distributed between neighboring bins using triangular
weights
$\max(0,1-d_{\pi}(\theta,\nu_k)/(\pi/9))$,
where $d_{\pi}$ is circular distance modulo $\pi$.
We pool the weighted votes over the patch and normalize by the
pooled gradient magnitude, with a small numerical stabilizer.
This yields nine values describing the relative prevalence
of different edge directions.
Our descriptor uses a single histogram per patch rather than
the full spatial cell and block construction of conventional HOG.

\paragraph{DCT energy: variation across spatial frequencies.}
We apply an orthonormal two-dimensional discrete cosine
transform (DCT)~\cite{ahmed1974discrete} to the grayscale patch.
Low-frequency coefficients describe slowly varying intensity
patterns, whereas higher-frequency coefficients describe
more rapid spatial changes.

We exclude the DC coefficient, which represents the patch mean,
and group the remaining coefficients into six radial frequency
bands. For band $\mathcal{F}_k$, we compute
\begin{equation}
  \phi_{\mathrm{dct}}^{k}(\tilde w_r)
  =
  \log\!\left(
    1+
    \frac{1}{|\mathcal{F}_k|}
    \sum_{(u,v)\in\mathcal{F}_k}C_{uv}^{2}
  \right),
  \label{eq:appendix_dct_energy}
\end{equation}
where $C_{uv}$ is a DCT coefficient.
Averaging accounts for different numbers of coefficients
per band, and the logarithm compresses the range of energy values.
The bands partition the radial coordinate $\sqrt{u^2+v^2}$ into six
equal-width intervals over $[0,\sqrt{2}(p-1)]$, with the maximum
radius assigned to the last band and the DC coefficient excluded.
This descriptor measures the distribution of spatial-frequency
energy rather than reconstructing the original patch.

\paragraph{LBP histogram: local intensity patterns.}
Local binary patterns (LBP) describe texture through comparisons
between a pixel and its neighbors~\cite{ojala1996comparative}.
For each grayscale pixel, we compare its intensity with those
of its eight neighbors:
\begin{equation}
  \operatorname{LBP}(a,b)
  =
  \sum_{t=0}^{7}2^t
  \mathbf{1}\!\left[I_t(a,b)\geq I(a,b)\right].
  \label{eq:appendix_lbp_code}
\end{equation}
where $I_t(a,b)$ denotes the intensity of neighbor $t$.
Neighbors are ordered clockwise from the top-left position
(top-left, top, top-right, right, bottom-right, bottom, bottom-left,
left). LBP codes are computed on the full grayscale image, with
replicate padding at the image boundary, before patch-wise pooling.

The resulting code lies between 0 and 255.
We coarsen these codes into 16 bins using
$q_{\mathrm{lbp}}=\lfloor\operatorname{LBP}/16\rfloor$
and normalize the histogram by the number of pixels in the patch.
This is a compact histogram of local intensity patterns;
it does not use uniform-pattern or rotation-invariant LBP mapping.

\paragraph{Raw-pixel PCA: dominant variation in RGB patches.}
The previous descriptors summarize predefined appearance
attributes. To include a more general description of pixel
variation, we also use principal component analysis
(PCA)~\cite{jolliffe2016principal} on flattened RGB patches.

Let $v_r=\operatorname{vec}(\tilde w_r)\in\mathbb{R}^{3p^2}$.
Using sampled probe-training patches, we estimate their mean
$\mu_{\mathrm{pca}}$ and the top 64 eigenvectors of their
covariance matrix, collected in
$U_{\mathrm{pca}}\in\mathbb{R}^{3p^2\times64}$.
The target is
\begin{equation}
  \phi_{\mathrm{raw\_pca}}(\tilde w_r)
  =
  U_{\mathrm{pca}}^\top(v_r-\mu_{\mathrm{pca}}).
  \label{eq:appendix_raw_pca}
\end{equation}
The mean and basis are then fixed and applied to both splits.
Unlike the pooled color and texture histograms, this descriptor
retains information about pixel positions through the flattened
patch representation. It captures the leading modes of RGB
variation rather than all raw-pixel information.

\subsection{Fitting the Linear Decoder}

For each backbone and evaluated layer, we fit a separate
ridge-regression decoder~\cite{hoerl1970ridge}.
Let $x_j\in\mathbb{R}^{d}$ be a frozen patch token and
$y_j\in\mathbb{R}^{122}$ its appearance target.
We standardize each feature dimension using its probe-training
mean $\mu_x$ and standard deviation $\sigma_x$, with a small
variance floor for numerical stability:
\begin{equation}
  \bar x_j=(x_j-\mu_x)\oslash\sigma_x,
  \qquad
  \bar y_j=y_j-\mu_y,
\end{equation}
where $\oslash$ denotes element-wise division and $\mu_y$ is
the probe-training target mean.
Targets are centered but are not variance-standardized.

Given $N$ sampled training pairs, the decoder is fitted by
\begin{equation}
  W^\star
  =
  \arg\min_W
  \frac{1}{N}\sum_{j=1}^{N}
  \left\|W^\top\bar x_j-\bar y_j\right\|_2^2
  +\lambda\|W\|_F^2.
  \label{eq:appendix_ridge_objective}
\end{equation}
The regularization term limits the magnitude of the decoder
weights. The implementation accumulates feature covariance
and feature--target cross-covariance statistics, then solves
\begin{equation}
  \left(\frac{\bar X^\top\bar X}{N}+\lambda I_d\right)W^\star
  =
  \frac{\bar X^\top\bar Y}{N},
  \label{eq:appendix_ridge_closed_form}
\end{equation}
where the rows of $\bar X$ and $\bar Y$ are the standardized
features and centered targets.
This avoids storing all sampled patch pairs simultaneously.
The parameter $\lambda$ corresponds to
\texttt{ridge\_alpha} in the implementation and is fixed at $10^{-3}$
for all reported probes.

For a validation token $x_j$, the prediction is
\begin{equation}
  \hat y_j
  =
  (W^\star)^\top
  \big((x_j-\mu_x)\oslash\sigma_x\big)+\mu_y.
  \label{eq:appendix_prediction}
\end{equation}
The backbone, preprocessing statistics, PCA basis, and decoder
are all fixed during validation.

\subsection{Scoring and Interpretation}

\paragraph{Dimension-averaged coefficient of determination.}
We evaluate each target dimension using the coefficient of
determination, $R^2$.
Let $\mathcal{V}$ denote the validation patch pairs and let
$\mu^{\mathrm{val}}_{g,k}$ be the validation mean of dimension
$k$ in descriptor group $g$.
We compute
\begin{equation}
  R^2_{g,k}
  =
  1-
  \frac{
    \sum_{j\in\mathcal{V}}
    (\hat y_{j,g,k}-y_{j,g,k})^2
  }{
    \max\!\left(
      \sum_{j\in\mathcal{V}}
      (y_{j,g,k}-\mu^{\mathrm{val}}_{g,k})^2,
      \epsilon
    \right)
  }.
  \label{eq:appendix_r2_dim}
\end{equation}
The validation mean is used only to compute the evaluation
denominator; it is not used to fit the decoder or make predictions.
We then average over the $m_g$ dimensions within each group:
\begin{equation}
  R^2_{g,\mathrm{mean}}
  =
  \frac{1}{m_g}\sum_{k=1}^{m_g}R^2_{g,k}.
  \label{eq:appendix_r2_mean}
\end{equation}
Computing $R^2$ before averaging gives each descriptor dimension
equal weight, rather than allowing dimensions with larger
target variance to dominate the group score.

\paragraph{Reading the scores.}
When the target's validation sum of squares exceeds $\epsilon$,
$R^2=1$ indicates perfect
prediction, $R^2=0$ matches the squared error of predicting
its validation mean, and $R^2<0$ indicates a larger error
than that baseline.
A higher group score therefore means that the corresponding
appearance attributes are more linearly accessible from the
patch tokens under this probe protocol.
The numerical floor $\epsilon$ stabilizes dimensions with
negligible validation variance.

\paragraph{What this diagnostic establishes.}
Local Appearance Decodability measures the linear accessibility
of selected pixel-derived attributes.
A lower score does not establish that the information has been
completely removed: it may remain accessible to a nonlinear
decoder or be encoded in a form not captured by these targets.
Likewise, a higher score does not imply better spatial matching.
A token can encode a patch's color or texture without preserving
the feature relationships needed to match that patch across views.
We therefore interpret this diagnostic together with
Local-to-Global Correspondence
(Section~\ref{sec:diag_local_global}), which evaluates a
complementary property of local representations.

\subsection{The Trajectories of Local Appearance Decodability}
\begin{figure}[H]
  \centering
  \includegraphics[width=12cm]{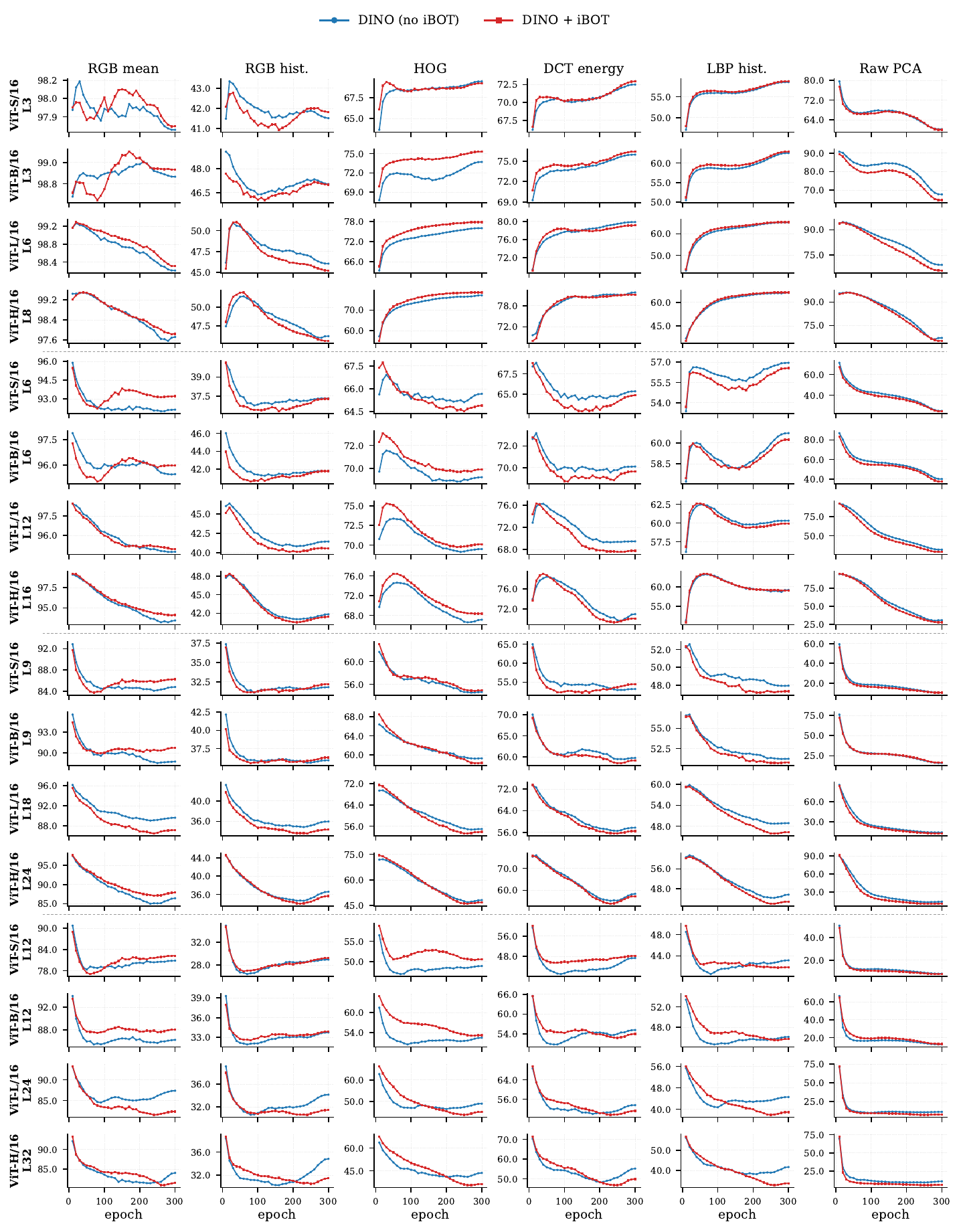}
  \caption{Training trajectories of local appearance decodability for each descriptor group.}
  \label{fig:decodability_trajectory}
\end{figure}

The decodability trajectories reveal a scale-dependent effect of iBOT
on local appearance information. For ViT-S/16 and ViT-B/16,
adding iBOT generally preserves or slightly improves descriptor
decodability, with no consistent loss of linearly recoverable
appearance evidence. For ViT-L/16 and ViT-H/16, reduced
final-layer decodability emerges during training and persists
at the final checkpoint across descriptor groups, including
RGB mean, DCT energy, HOG, and raw PCA.
Together with the layer-wise results, these trajectories suggest
that the appearance cost of iBOT is concentrated in the deeper
layers of larger backbones, rather than being a uniform
consequence of patch self-distillation at every model scale.

\subsection{Visualization of Local Appearance Decodability}

RGB-mean predictions are displayed directly as patch colors, clipped
to $[0,1]$. For HOG, DCT energy, and raw PCA, we visualize the
$\ell_2$ norm of each patch's descriptor vector as a heatmap.
Each heatmap is independently min--max normalized; colors therefore
show relative spatial patterns within a panel, not calibrated
prediction magnitudes across models. These maps do not reconstruct
RGB pixels or display the individual descriptor components.
Quantitative comparisons use the full descriptor vectors and $R^2$.

\begin{figure}[H]
  \centering
  \includegraphics[width=12cm]{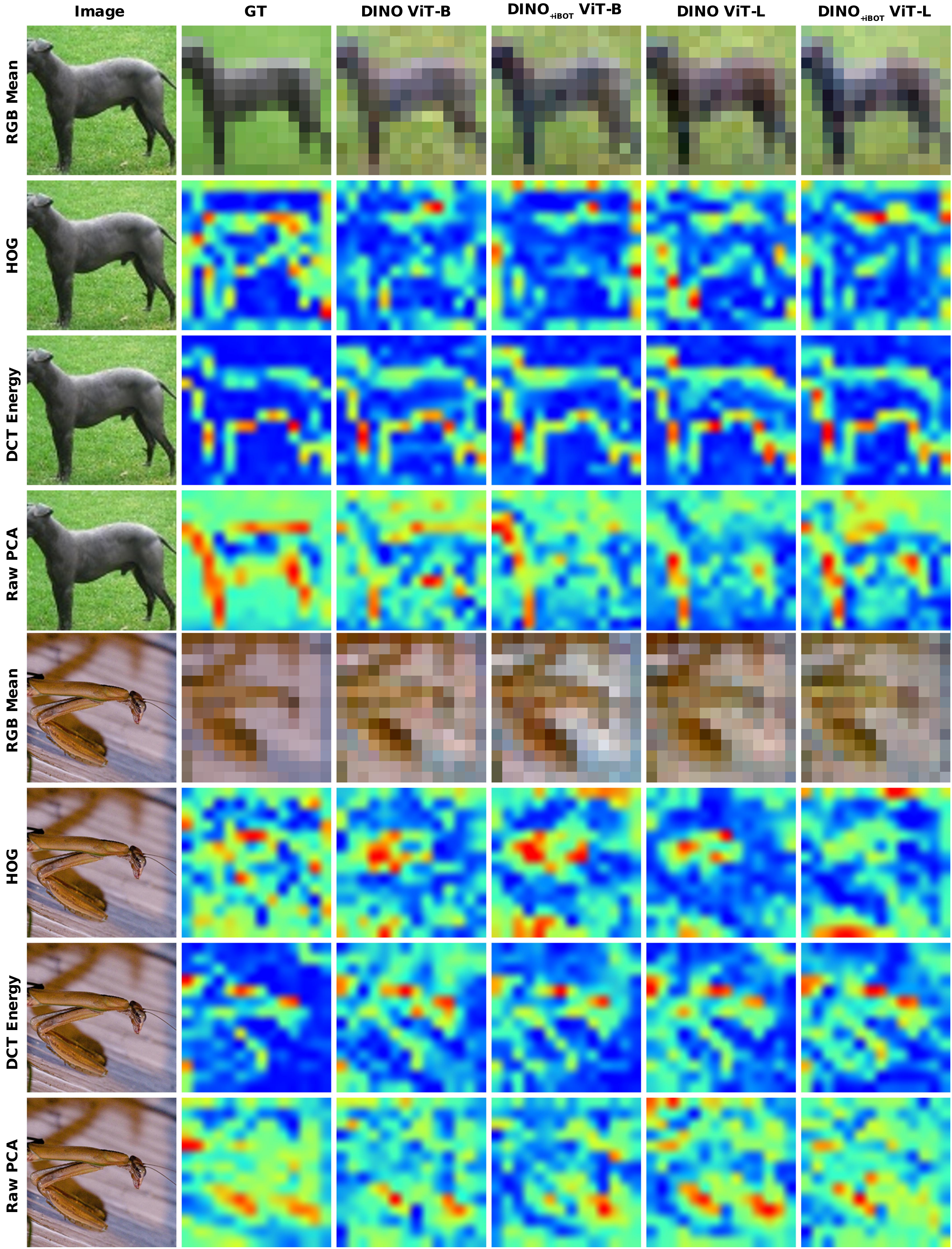}
  \caption{Visualization of local appearance decodability. Here we compare the RGB mean, HOG, DCT energy, and raw PCA results produced by ViT-B/16 and ViT-L/16 models of DINO and DINO$_{+\mathrm{iBOT}}$.}
  \label{fig:local_decode_visual}
\end{figure}

\section{Implementation and Visualization of Local-to-Global Correspondence}
\label{app:local_global_details}

Local-to-Global Correspondence asks whether a representation can
locate a cropped image region within its source image.
We extract frozen patch tokens from the crop and the full image,
match them using cosine similarity, and evaluate whether the
matches identify the correct source region and support a
consistent geometric localization.
The reference region is known from the crop operation, so the
diagnostic requires no human annotations or learned matching head.

\subsection{Constructing Local Queries}

\paragraph{Source images and paired evaluation.}
We use the 157 source images from the Copydays
dataset~\cite{douze2009evaluation}.
For each image, we sample four local crops, giving 628 queries
per run. The default crop area ratio is $\rho=0.1$.
The crop specifications are cached and reused across models
in each paired comparison, ensuring that DINO and
DINO$_{+\mathrm{iBOT}}$ are evaluated on identical queries.

\paragraph{Crop size and position.}
For a source image of width $W_{\mathrm{img}}$ and height
$H_{\mathrm{img}}$, the target crop area is
$A_{\mathrm{crop}}=\rho W_{\mathrm{img}}H_{\mathrm{img}}$.
To introduce variation in crop shape, we sample the aspect ratio
$\alpha$ log-uniformly from $[0.75,1.25]$.
The crop dimensions are
\begin{equation}
  W_{\mathrm{crop}}
  =
  \operatorname{round}\!\sqrt{A_{\mathrm{crop}}\alpha},
  \qquad
  H_{\mathrm{crop}}
  =
  \operatorname{round}\!\sqrt{A_{\mathrm{crop}}/\alpha}.
  \label{eq:app_crop_size}
\end{equation}
We clip these dimensions to the valid image range and sample the
top-left corner $(x_0,y_0)$ uniformly from valid integer positions.
The resulting crop occupies approximately a fraction $\rho$
of the source image, subject to rounding and clipping.

We express its reference box in normalized source-image coordinates:
\begin{equation}
  B^\star
  =
  \left(
    \frac{x_0}{W_{\mathrm{img}}},
    \frac{y_0}{H_{\mathrm{img}}},
    \frac{x_0+W_{\mathrm{crop}}}{W_{\mathrm{img}}},
    \frac{y_0+H_{\mathrm{crop}}}{H_{\mathrm{img}}}
  \right)
  =
  (x_0^\star,y_0^\star,x_1^\star,y_1^\star).
  \label{eq:app_crop_box_norm}
\end{equation}
Both the full image and the crop are resized to $224\times224$
before feature extraction. Consequently, the same visual content
appears at different scales and with different surrounding
context in the two inputs.

\subsection{Computing Patch Similarities}

For a fixed backbone and layer $\ell$, let $\mathcal{I}$ and
$\mathcal{J}$ index the local-query and source-image patch tokens,
respectively. Both inputs produce an
$H_{\mathrm{grid}}\times W_{\mathrm{grid}}$ patch grid.
Let $u(i)\in[0,1]^2$ denote the normalized center of query patch
$i$, and let $r(j)\in[0,1]^2$ denote the normalized center of
source patch $j$.

\paragraph{Reference source region.}
The reference region consists of source patches whose centers
fall inside the sampled crop:
\begin{equation}
  G(B^\star)
  =
  \left\{
    j\in\mathcal{J}:
    x_0^\star\le r_x(j)<x_1^\star,\;
    y_0^\star\le r_y(j)<y_1^\star
  \right\}.
  \label{eq:app_local_global_gt}
\end{equation}
If the box contains no source-patch center, we use the source
patch nearest to the box center as a fallback.

\paragraph{Similarity matrix.}
We extract patch tokens from the frozen backbone and normalize
each token to unit $\ell_2$ norm.
The local-to-global similarity matrix is
\begin{equation}
  A_\ell(i,j)
  =
  \tilde z_\ell^{\mathrm{local}}(i)^\top
  \tilde z_\ell^{\mathrm{global}}(j),
  \qquad
  i\in\mathcal{I},\ j\in\mathcal{J},
  \label{eq:app_local_global_similarity}
\end{equation}
where $\tilde z$ denotes a normalized token.
For each query patch, its top-1 source match is
\begin{equation}
  m(i)=\operatorname*{arg\,max}_{j\in\mathcal{J}}A_\ell(i,j).
  \label{eq:app_patch_top1_match}
\end{equation}
All matching and localization metrics below are computed from
this similarity matrix.

\subsection{Matching and Localization Metrics}

We evaluate complementary aspects of correspondence:
whether individual matches fall inside the correct region,
whether the strongest responses recover that region, and
whether the matches jointly support the correct crop location.

\paragraph{Patch matching accuracy: retrieving the correct region.}
We measure the fraction of query patches whose top-1 match
lies inside the reference source region:
\begin{equation}
  T_{\mathrm{patch}}^{1}(\ell)
  =
  \frac{1}{|\mathcal{I}|}
  \sum_{i\in\mathcal{I}}
  \mathbf{1}\!\left[m(i)\in G(B^\star)\right].
  \label{eq:app_local_global_patch_top1}
\end{equation}
A higher score indicates that query patches preferentially
match their region of origin rather than unrelated parts
of the source image.

This is a region-membership criterion: a match is counted as
correct anywhere inside $G(B^\star)$.
It does not require the source patch to coincide with the
exact geometrically corresponding position.
We therefore interpret this score together with the geometric
localization metrics below.

For completeness, the top-$k$ version counts a query patch as
successful if at least one of its $k$ highest-similarity source
patches belongs to $G(B^\star)$:
\begin{equation}
  T_{\mathrm{patch}}^{k}(\ell)
  =
  \frac{1}{|\mathcal{I}|}
  \sum_{i\in\mathcal{I}}
  \mathbf{1}\!\left[
    M_k(i)\cap G(B^\star)\neq\emptyset
  \right],
  \label{eq:app_local_global_patch_topk}
\end{equation}
where $M_k(i)$ is the set of top-$k$ matches.
Our reported patch matching accuracy uses $k=1$.

\paragraph{Region IoU: recovering the spatial extent of the crop.}
To obtain a response map over the source image, we assign
each source patch its highest similarity to any query patch:
\begin{equation}
  s(j)=\max_{i\in\mathcal{I}}A_\ell(i,j).
  \label{eq:app_source_response}
\end{equation}
Thus, a source patch receives a high response when it strongly
matches at least one patch in the query.

Let $n_G=|G(B^\star)|$.
We select the $n_G$ source patches with the highest response
scores and denote this predicted patch set by $R(B^\star)$.
Its overlap with the reference region is
\begin{equation}
  T_{\mathrm{region}}(\ell)
  =
  \frac{|R(B^\star)\cap G(B^\star)|}
       {|R(B^\star)\cup G(B^\star)|}.
  \label{eq:app_local_global_region_iou}
\end{equation}
A high score means that the strongest responses concentrate
within the correct crop region.

The reference region size determines how many patches are
selected, but not which patches are selected.
This avoids choosing a similarity threshold and evaluates
localization at a fixed region size.
It does not make the task independent of crop ratio:
larger crops occupy more of the source grid and can differ
in matching difficulty.

\paragraph{Geometric localization: agreement on the crop position.}
We next test whether the top-1 matches support a consistent
location for the query within the source image.
This evaluation uses the known normalized crop dimensions
\[
  w^\star=x_1^\star-x_0^\star,
  \qquad
  h^\star=y_1^\star-y_0^\star.
\]
For a geometrically correct match, the source position should
approximately satisfy
\[
  r(m(i))
  \approx
  (x_0^\star,y_0^\star)
  +(w^\star,h^\star)\odot u(i),
\]
where $\odot$ denotes element-wise multiplication.
Each query patch therefore casts a vote for the crop's
top-left corner:
\begin{equation}
  t_i
  =
  r(m(i))-(w^\star,h^\star)\odot u(i).
  \label{eq:app_local_global_vote}
\end{equation}
Correct matches should produce similar votes, whereas
inconsistent matches generally produce dispersed votes.

We select the location supported by the largest consensus
set, following the robust-consensus principle of
RANSAC~\cite{fischler1981random}.
Specifically, we evaluate at most 128 candidate votes,
selected at uniformly spaced indices in the vote list.
For each candidate, we count votes within an $\ell_\infty$
distance
\begin{equation}
  \tau
  =
  \frac{1.5}{\min(H_{\mathrm{grid}},W_{\mathrm{grid}})}.
  \label{eq:app_ransac_tolerance}
\end{equation}
The candidate with the largest count is selected as
$\hat t=(\hat t_x,\hat t_y)$.
This is a RANSAC-style translation consensus with known crop
dimensions, rather than an unconstrained affine or homography
estimation procedure.

The predicted box is
\begin{equation}
  \hat B
  =
  (\hat t_x,\hat t_y,
   \hat t_x+w^\star,\hat t_y+h^\star),
  \label{eq:app_pred_box}
\end{equation}
with coordinates clipped to $[0,1]$.
We report its intersection-over-union with the reference box:
\begin{equation}
  T_{\mathrm{box}}(\ell)
  =
  \operatorname{IoU}(\hat B,B^\star).
  \label{eq:app_local_global_box_iou}
\end{equation}
We also report the IoU$_{50}$ success rate, obtained by averaging
\begin{equation}
  T_{\mathrm{succ}}(\ell)
  =
  \mathbf{1}\!\left[T_{\mathrm{box}}(\ell)\ge0.5\right]
  \label{eq:app_local_global_success}
\end{equation}
over queries.
Box IoU measures localization quality continuously, whereas
IoU$_{50}$ measures how often localization reaches a fixed
overlap threshold.

\paragraph{Auxiliary inlier ratio.}
The fraction of votes supporting the selected location is
\begin{equation}
  T_{\mathrm{inlier}}(\ell)
  =
  \frac{1}{|\mathcal{I}|}
  \sum_{i\in\mathcal{I}}
  \mathbf{1}\!\left[
    \|t_i-\hat t\|_\infty\le\tau
  \right].
  \label{eq:app_local_global_inlier}
\end{equation}
This measures internal geometric agreement among matches.
A consistent set of incorrect matches can also produce a
high inlier ratio, so this quantity should be interpreted
alongside box IoU and localization success.

\paragraph{Aggregation and interpretation.}
We compute each metric separately for each query and then
average over the 628 crops.
Higher patch matching accuracy, region IoU, box IoU, and
IoU$_{50}$ indicate better performance under their respective
criteria.

Together, these metrics assess cross-view region retrieval
and geometric consistency under cropping and resizing.
They do not measure whether pixel values or texture
descriptors can be reconstructed from the tokens, nor do
they establish robustness to arbitrary viewpoint changes.
We therefore use this diagnostic alongside Local Appearance
Decodability to distinguish spatial correspondence from
recoverable appearance evidence.

\vspace{-0.2in}
\begin{figure}[H]
  \centering
  \includegraphics[width=12cm]{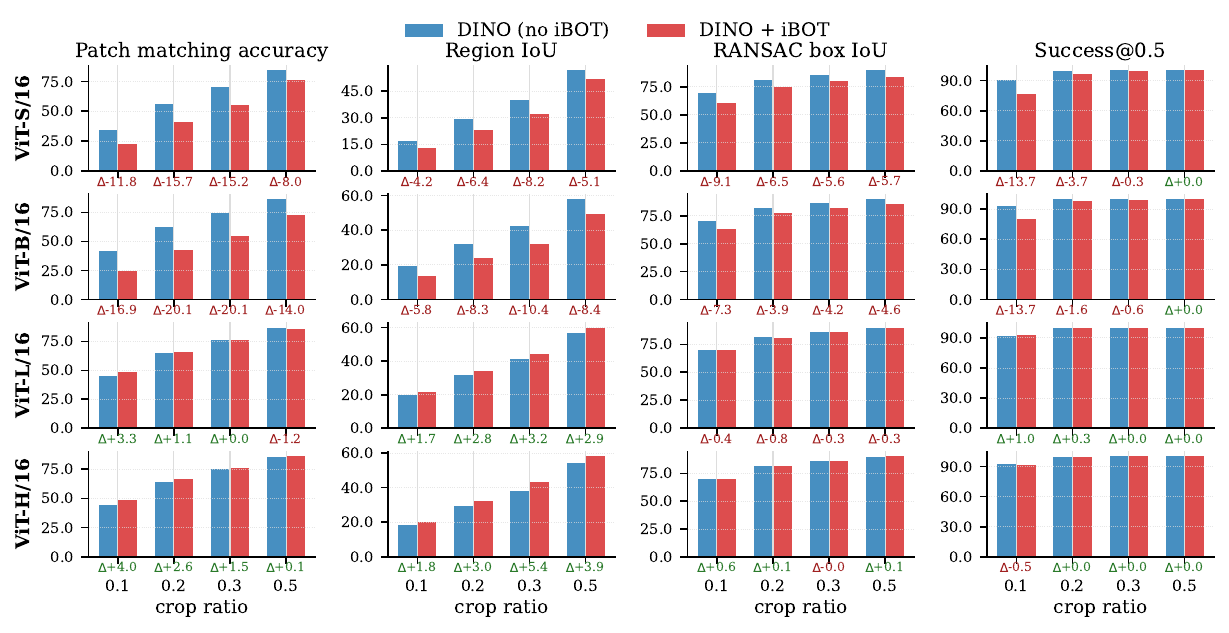}
  \vspace{-0.15in}
  \caption{
  Local-to-global correspondence across crop area ratios.
  Each ratio specifies the approximate fraction of the source
  image used to construct a local query.
  $\Delta$ denotes DINO$_{+\mathrm{iBOT}}$ minus DINO under
  matched pretraining settings and identical sampled crops.
  Positive values favor DINO$_{+\mathrm{iBOT}}$ for all
  reported metrics.
  }
  \label{fig:crop_copy_ratio}
\end{figure}

\vspace{-0.2in}
\subsection{Crop Ratios and Sampling}

Figure~\ref{fig:crop_copy_ratio} evaluates whether the paired
differences persist when the query covers different fractions
of the source image.
Changing the crop ratio changes both the amount of available
visual context and the scale difference between the resized
query and source image.
We therefore emphasize comparisons between models at the
same crop ratio.

All reported correspondence results use a single query-sampling run
with seed 0. The same sampled queries are used for each paired
comparison. We do not estimate variability across sampling seeds
or independent pretraining runs; the plots show point estimates,
and small differences should not be interpreted as statistically
established gains.

\subsection{Visualization and Training Trajectories}
\begin{figure}[H]
  \centering
  \includegraphics[width=12cm]{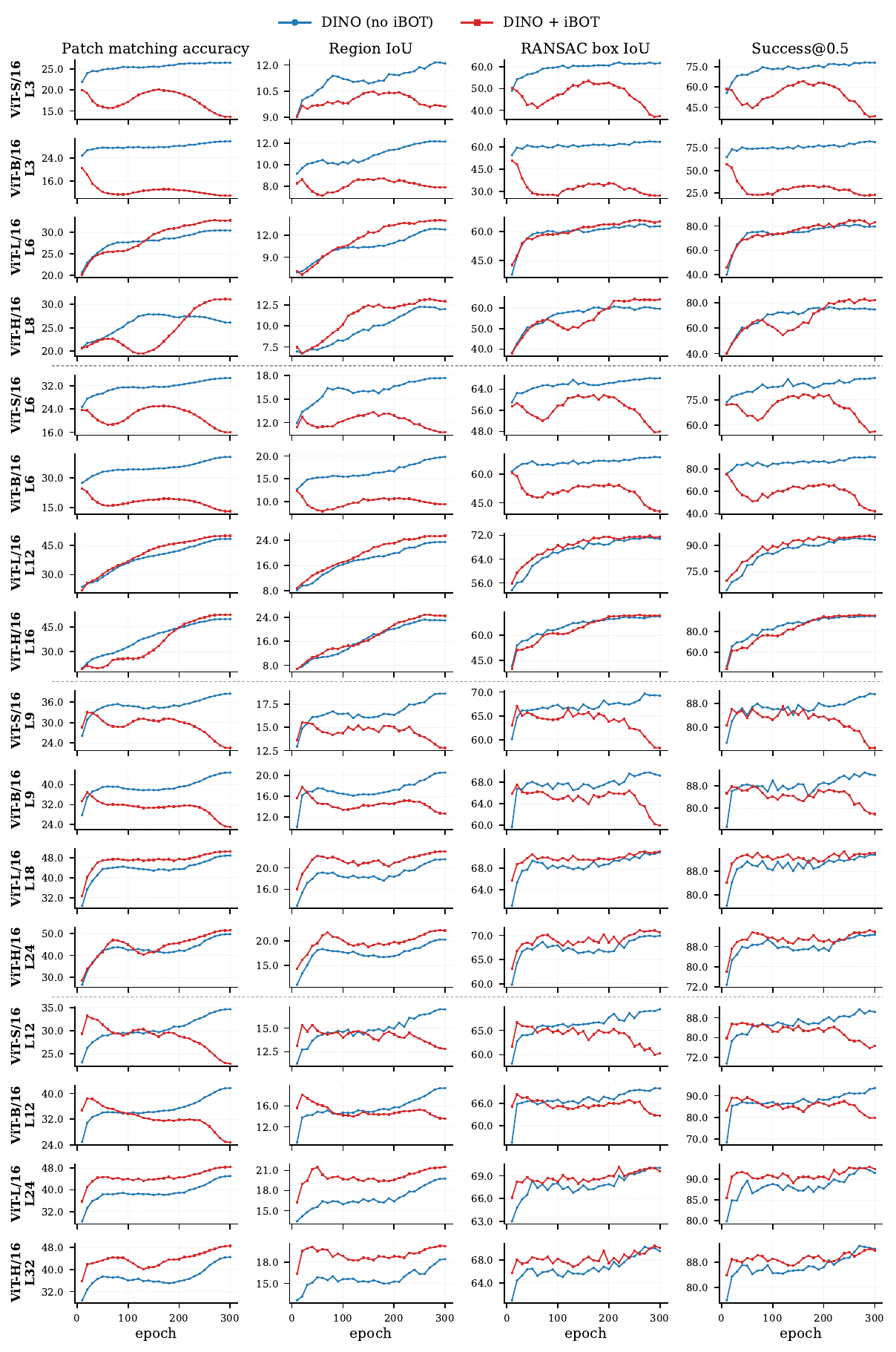}
  \caption{
  Training trajectories of local-to-global correspondence
  for DINO and DINO$_{+\mathrm{iBOT}}$ across backbone scales.
  Rows show the backbone and evaluated layer: blocks 3/6/9/12 for
  ViT-S/B, 6/12/18/24 for ViT-L, and 8/16/24/32 for ViT-H.
  The sustained deterioration observed in smaller
  $+$iBOT models is alleviated in larger backbones.
  }
  \label{fig:crop_copy_trajectory}
\end{figure}
\begin{figure}[H]
  \centering
  \includegraphics[width=12cm]{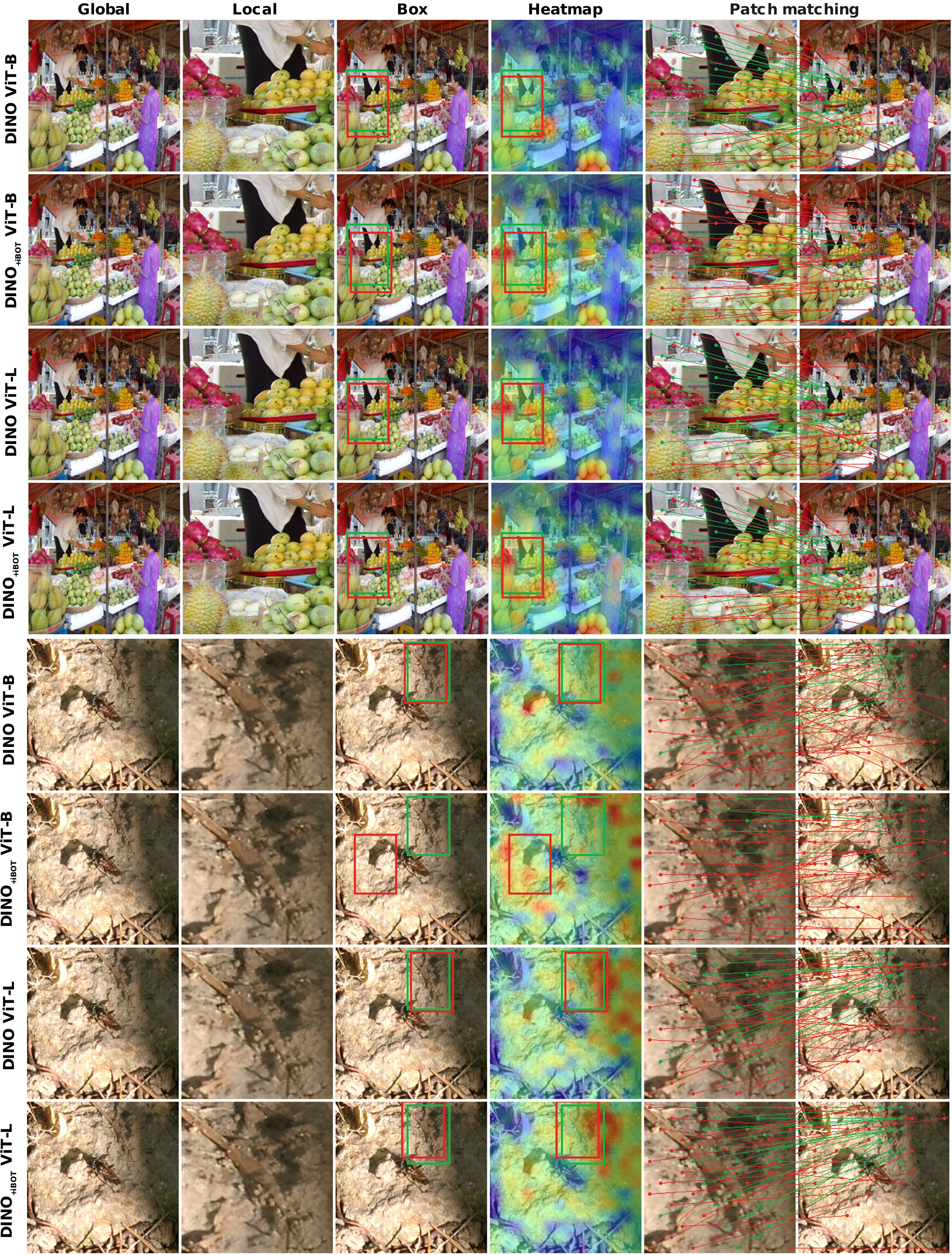}
  \caption{Additional visualization of local-global correspondence for ViT-B/16 and ViT-L/16 models of DINO and DINO$_{+\mathrm{iBOT}}$.}
  \label{fig:local_global_extra}
\end{figure}

\paragraph{Reading the qualitative examples.}
The qualitative results in Figure~\ref{fig:local_global_extra}
show complementary aspects of the same matching process.
Similarity heatmaps indicate where the query produces strong
responses in the source image, patch matches show the
individual associations, and estimated boxes summarize
their geometric consensus.
These views help distinguish broadly distributed responses,
incorrect individual matches, and failures to localize the
query as a coherent region.

Figure~\ref{fig:crop_copy_trajectory} shows that the effect of
iBOT depends on model scale.
For ViT-S/16 and ViT-B/16, final-layer patch matching accuracy
declines after an early peak under $+$iBOT training, while
DINO generally improves or plateaus.
Related deterioration appears in region-level and geometric
localization metrics, although its magnitude varies.
For ViT-L/16 and ViT-H/16, this sustained decline is no longer
evident, and adding iBOT generally preserves or modestly
improves correspondence, consistent with the DAVIS results.

These trajectories suggest that the correspondence cost
observed in smaller models develops during optimization
and can be alleviated at larger scales.
However, this improvement coexists with reduced deep-layer
appearance decodability in larger backbones.
Thus, improved matching does not imply that pixel and texture
evidence also becomes more accessible: the two diagnostics
measure distinct properties of the learned representations.

\subsection{Additional Visualization of Local-Global Correspondence}

\clearpage

\clearpage
\section{Extended Related Work}
\label{app:related_work}

We organize the related literature around three questions:
how self-distillation supervises patch representations,
how prediction targets shape the information being learned,
and how relational regularization changes these properties.

\subsection{From Image-Level to Patch-Level Self-Distillation}
\label{app:rw_joint_embedding}

DINO~\citep{caron2021dino} learns visual representations by
matching teacher and student predictions across augmented views.
Despite using an image-level objective, it exhibits emergent
spatial and semantic structure in its ViT features.
iBOT~\citep{zhou2022ibot} extends this approach with masked
patch self-distillation: student tokens at masked locations
predict the teacher's patch-level target distributions,
alongside class-token distillation.
DINOv2~\citep{oquab2023dinov2} incorporates these image-level
and patch-level objectives into a scalable training recipe,
and DINOv3~\citep{simeoni2025dinov3} retains them while further
scaling visual pretraining.

These developments establish the effectiveness of combining
global and patch-level supervision for transferable visual
representations. Our work examines the contribution of the
patch objective through paired DINO and
DINO$_{+\mathrm{iBOT}}$ experiments.
The resulting semantic gains coexist with different local
effects across model scales: correspondence deteriorates in
smaller backbones, whereas larger backbones preserve or improve
correspondence while exposing less linearly recoverable
appearance evidence in deep layers.
Thus, the spatial granularity of supervision alone does not
determine which local properties are preserved.

\subsection{Prediction Targets in Masked Image Modeling}
\label{app:rw_mim}

Masked image modeling methods differ substantially in what
they require the model to predict.
MAE~\citep{2022_cvpr_He} and SimMIM~\citep{xie2022simmim}
use pixel reconstruction, while
MaskFeat~\citep{wei2022maskfeat} studies feature targets and
finds HOG descriptors effective for visual pretraining.
These objectives directly supervise aspects of image
appearance.
Other approaches predict learned targets:
BEiT~\citep{bao2021beit} uses discrete visual tokens,
data2vec~\citep{baevski2022data2vec} predicts contextualized
teacher representations, and
I-JEPA~\citep{2023_cvpr_Assran} predicts target-block
representations from a context block.
iBOT similarly uses learned teacher targets, with an online
tokenizer trained jointly with the student.

This distinction motivates our analysis of appearance
decodability. Predicting a learned representation need not
require retaining every pixel-derived attribute of the
corresponding patch.
However, the target type alone does not establish what remains
accessible in the resulting features.
Our experiments show that adding iBOT leaves appearance
decodability largely unchanged or slightly improved in smaller
models, while reducing it in deeper layers of larger models.
The consequences of semantic patch supervision therefore
depend on model scale and depth.
Our diagnostic uses appearance descriptors as evaluation
targets for frozen features; it does not introduce an
additional reconstruction objective during pretraining.

\subsection{Gram Anchoring and the Preservation of Patch Relations}
\label{app:rw_gram}

DINOv3~\citep{simeoni2025dinov3} observes degradation in dense
feature quality during extended training and introduces Gram
anchoring to regularize pairwise patch similarities.
The objective encourages the student's patch Gram matrix
to match that of an earlier teacher checkpoint, preserving
relational structure while allowing the feature vectors
themselves to change.

Our experiments examine this intervention through the
distinction between appearance and correspondence.
We compare standard Gram anchoring with a variant initialized
from a DINO checkpoint.
In ViT-B, the DINO-based variant substantially improves
correspondence, with limited and mixed changes in appearance
decodability.
In ViT-L, it improves deep-layer appearance decodability
without consistently improving correspondence.
These observations suggest that the effects of anchoring
depend on both model scale and the reference representations.

This extends the analysis of relational regularization beyond
aggregate dense-task performance.
Matching within-view patch relations does not explicitly
constrain cross-view consistency or the recoverability of
pixel-derived attributes.
Our results show that these properties need to be evaluated
separately when assessing whether an intervention preserves
local visual information.

\section{Limitations}
\label{sec:limitations}

Our study has several limitations.
First, computational constraints restrict pretraining to
ImageNet-1K for 300 epochs. Although we examine multiple
backbone scales, this setting is substantially smaller than
the data and compute regimes of DINOv2 and DINOv3.
Whether the observed trade-offs persist under larger datasets
and longer training remains to be established.

Second, we fix the iBOT loss weight to $1$, following the
standard weighting used in DINOv2 and DINOv3.
Other weights or schedules may change the balance between
semantic learning, appearance decodability, and correspondence.
Our results therefore characterize this commonly used setting
rather than an unavoidable consequence of masked patch
self-distillation.

Third, our diagnostics measure specific aspects of local
representations. Linear decodability does not capture
information accessible only through nonlinear mappings,
and crop-based correspondence does not cover the full range
of viewpoint changes, deformations, and occlusions encountered
in real-world matching.
The region-membership score does not test exact pointwise matching,
and geometric localization uses known crop dimensions.
We also use single pretraining and query-sampling runs, so small
differences are descriptive rather than statistically established.
Our optical-flow results evaluate a learned head on the same Sintel
videos used to fit it and do not measure held-out generalization.
Finally, the observed dependence on model scale and Gram
anchoring motivates possible explanations, but does not
establish their underlying mechanisms.
Further controlled experiments are needed to distinguish
the roles of model capacity, optimization, and teacher
representations. In particular, drop-path rates differ between
ViT-S/B and ViT-L/H, and ViT-H uses a smaller batch size and scaled
learning rate. The paired comparisons are controlled within each
backbone, but cross-scale differences cannot be attributed to model
capacity alone. Likewise, enabling iBOT also enables student masking;
we do not separately isolate the effects of masking and its loss.

\section{Licenses for Existing Assets}
\label{app:asset_licenses}

This work uses existing datasets, benchmark annotations, model architectures,
and public research code. We cite the corresponding papers throughout the main
text and appendix, and use each asset only for research benchmarking under the
terms stated by its original provider. The planned public release will contain our
code, scripts, generated measurements, and derived metadata needed for
reproduction; it will not redistribute raw third-party images or videos unless
their original terms explicitly permit redistribution.


\begin{table}[H]
  \centering
  \footnotesize
  \renewcommand{\arraystretch}{1.0}
  \caption{Datasets and benchmarks used in the paper.}
  \label{tab:dataset_licenses}
  \begin{tabularx}{\linewidth}{@{} L{1.9cm} L{2.7cm} L{3.0cm} Y @{}}
    \toprule
    \textbf{Asset} & \textbf{Use in this paper} & \textbf{Credit / source} & \textbf{License or terms of use} \\
    \midrule
    ImageNet-1K
      & Self-supervised pretraining, $k$-NN evaluation, and diagnostic image splits
      & \citet{2009_ImageNet}; \href{https://www.image-net.org/download.php}{official access terms}
      & Access provided for non-commercial research and educational use under ImageNet terms. ImageNet does not own the underlying image copyrights, which remain with the original owners. \\
    \addlinespace
    PASCAL VOC \mbox{2007 / 2012}
      & VOC2007 object detection and VOC2012 semantic segmentation
      & \citet{2010_IJCV_VOC,everingham2015pascal}; \href{https://www.robots.ox.ac.uk/\%7Evgg/projects/pascal/VOC/}{VOC challenge pages}
      & The official database-rights statement notes that VOC contains Flickr images, whose use must respect the corresponding Flickr terms of use. \\
    \addlinespace
    MS COCO
      & Copy-detection distractors and whitening images
      & \citet{2014_ECCV_COCO}; \href{https://cocodataset.org/\#termsofuse}{COCO terms of use}
      & COCO annotations are released under CC BY 4.0; COCO images are subject to the Flickr terms and to the image-level license metadata provided by COCO. \\
    \addlinespace
    $\mathcal{R}$Oxford5k / $\mathcal{R}$Paris6k
      & Image retrieval evaluation
      & \citet{radenovic2018revisiting}; \href{https://cmp.felk.cvut.cz/revisitop/}{Revisited Oxford / Paris benchmark}
      & Public benchmark data and annotations used for academic evaluation. The project page does not list a standard open-source license for the images; we do not redistribute the raw images. \\
    \addlinespace
    ADE20K
      & Similarity heatmaps and reference semantic masks
      & \citet{zhou2017scene}; \href{https://ade20k.csail.mit.edu/terms/}{official terms of use}
      & Images are provided for non-commercial research and education; copyright remains with their owners. The official terms specify BSD-3 terms for the annotations and software. \\
    \addlinespace
    INRIA Copydays
      & Copy detection and local-to-global correspondence diagnostics
      & \citet{douze2009evaluation}; \href{https://thoth.inrialpes.fr/\%7Ejegou/data.php.html}{Copydays project page}
      & INRIA holds copyright on the dataset images and distributes them ``as is'' without express or implied warranties. Used here only as a research benchmark; raw images are not redistributed. \\
    \addlinespace
    DAVIS-2017
      & Video object tracking evaluation
      & \citet{ponttuset2017davis}; \href{https://github.com/davisvideochallenge/davis-2017}{DAVIS 2017 repository}
      & Released under the BSD license, per the official repository. \\
    \addlinespace
    MPI Sintel
      & Optical flow evaluation
      & \citet{butler2012naturalistic}; \href{https://is.mpg.de/en/code/sintel-optical-flow-dataset}{MPI-IS dataset page}
      & Released under PS:License 1.0, with copyright held by the Max-Planck-Gesellschaft, per the official MPI-IS page. \\
    \bottomrule
  \end{tabularx}
\end{table}

\begin{table}[H]
  \centering
  \footnotesize
  \renewcommand{\arraystretch}{1.0}
  \caption{Existing codebases, model releases, and software assets used in the paper.}
  \label{tab:software_licenses}
  \begin{tabularx}{\linewidth}{@{} L{2.0cm} L{3.0cm} L{3.0cm} Y @{}}
    \toprule
    \textbf{Asset} & \textbf{Use in this paper} & \textbf{Credit / source} & \textbf{License or terms of use} \\
    \midrule
    DINOv2
      & Pretraining framework and ViT implementation
      & \citet{oquab2023dinov2}; \href{https://github.com/facebookresearch/dinov2}{DINOv2 repository}
      & Apache License 2.0 for the DINOv2 code and standard model weights, per the repository license. \\
    \addlinespace
    MMSegmentation
      & Linear semantic segmentation evaluation
      & \href{https://github.com/open-mmlab/mmsegmentation}{MMSegmentation repository}
      & Apache License 2.0. \\
    \addlinespace
    MMDetection
      & Frozen-backbone object detection evaluation
      & \href{https://github.com/open-mmlab/mmdetection}{MMDetection repository}
      & Apache License 2.0. \\
    \addlinespace
    MMFlow
      & Frozen-backbone optical-flow evaluation and PWCNet implementation
      & \href{https://github.com/open-mmlab/mmflow}{MMFlow repository}; PWCNet paper \citep{sun2018pwc}
      & Apache License 2.0 for the MMFlow toolbox. \\
    \addlinespace
    MAE
      & Raw-pixel masked-modeling baseline for local-appearance decodability
      & \citet{2022_cvpr_He}; \href{https://github.com/facebookresearch/mae}{MAE repository}
      & CC BY-NC 4.0, per the repository license. \\
    \addlinespace
    SimMIM
      & Raw-pixel masked-modeling baseline for local-appearance decodability
      & \citet{xie2022simmim}; \href{https://github.com/microsoft/SimMIM}{SimMIM repository}
      & MIT License, per the repository license. \\
    \bottomrule
  \end{tabularx}
\end{table}

\section*{NeurIPS Paper Checklist}

\begin{enumerate}

\item {\bf Claims}
    \item[] Question: Do the main claims made in the abstract and introduction accurately reflect the paper's contributions and scope?
    \item[] Answer: \answerYes{}.
    \item[] Justification: The abstract and introduction describe a controlled empirical
    study of the scale-dependent effects of masked patch
    self-distillation. The claims concern downstream performance,
    the distinction between local appearance decodability and
    local-to-global correspondence, and the effects of Gram
    anchoring. These claims are supported by
    Sections~\ref{sec:phenomenon}, \ref{sec:mechanism},
    and~\ref{sec:gram}. Possible mechanisms are presented as
    hypotheses rather than established explanations.

\item {\bf Limitations}
    \item[] Question: Does the paper discuss the limitations of the work performed by the authors?
    \item[] Answer: \answerYes{}.
    \item[] Justification:  Section~\ref{sec:limitations} discusses the restricted
    pretraining data and compute scale, the fixed iBOT loss
    weight, the scope of the two diagnostics, and the absence
    of a conclusive mechanistic explanation for the observed
    effects. The conclusions are limited to the evaluated
    models and training settings.

\item {\bf Theory assumptions and proofs}
    \item[] Question: For each theoretical result, does the paper provide the full set of assumptions and a complete (and correct) proof?
    \item[] Answer: \answerNA{}.
    \item[] Justification:
    The paper presents empirical analyses and diagnostic
    definitions. It does not introduce theoretical results
    requiring formal proofs.

    \item {\bf Experimental result reproducibility}
    \item[] Question: Does the paper fully disclose all the information needed to reproduce the main experimental results of the paper to the extent that it affects the main claims and/or conclusions of the paper (regardless of whether the code and data are provided or not)?
    \item[] Answer: \answerYes{}.
    \item[] Justification: Appendix~\ref{app:experimental_settings} describes the pretraining setup, model architectures, optimization hyperparameters, data, and downstream evaluation protocols. Appendices~\ref{app:descriptor_decodability} and~\ref{app:local_global_details} provide the diagnostic definitions and implementation details needed to reproduce the annotation-free analyses.

\item {\bf Open access to data and code}
    \item[] Question: Does the paper provide open access to the data and code, with sufficient instructions to faithfully reproduce the main experimental results, as described in supplemental material?
    \item[] Answer: \answerNo{}.
    \item[] Justification: A complete public code release is planned, but an
    accessible release is not yet provided with this version. The release will
    include pretraining and evaluation scripts, configuration files, environment
    setup, and data-access instructions. The appendix documents the experimental
    protocols and cites the public datasets and codebases used.

\item {\bf Experimental setting/details}
    \item[] Question: Does the paper specify all the training and test details (e.g., data splits, hyperparameters, how they were chosen, type of optimizer) necessary to understand the results?
    \item[] Answer: \answerYes{}.
    \item[] Justification: Appendix~\ref{app:experimental_settings} specifies the pretraining objective variants, backbone details, optimizer, schedules, crop settings, and all downstream evaluation protocols. Appendices~\ref{app:descriptor_decodability} and~\ref{app:local_global_details} further specify the diagnostic splits, sampling, probe fitting, and metrics.

\item {\bf Experiment statistical significance}
    \item[] Question: Does the paper report error bars suitably and correctly defined or other appropriate information about the statistical significance of the experiments?
    \item[] Answer: \answerNo{}.
    \item[] Justification: The main comparisons use single pretraining runs,
    and the correspondence diagnostic uses a single sampling seed (0).
    We do not report repeated-run standard deviations, confidence
    intervals, or significance tests. Appendices~\ref{app:local_global_details}
    and~\ref{sec:limitations} state these limits; small differences
    are interpreted descriptively.

\item {\bf Experiments compute resources}
    \item[] Question: For each experiment, does the paper provide sufficient information on the computer resources (type of compute workers, memory, time of execution) needed to reproduce the experiments?
    \item[] Answer: \answerYes{}.
    \item[] Justification: Appendix~\ref{app:experimental_settings} reports the hardware setup, mixed-precision and sharding strategy, and Table~\ref{tab:compute_resources_pretraining} reports approximate training times
    and memory requirements for the 300-epoch DINO and
    DINO$_{+\mathrm{iBOT}}$ runs across ViT-S/16, ViT-B/16,
    ViT-L/16, and ViT-H/16.
    
\item {\bf Code of ethics}
    \item[] Question: Does the research conducted in the paper conform, in every respect, with the NeurIPS Code of Ethics \url{https://neurips.cc/public/EthicsGuidelines}?
    \item[] Answer: \answerYes{}.
    \item[] Justification: The work is a self-supervised vision representation analysis using standard public datasets and benchmark protocols, with no new human-subject data collection or deployment of a decision-making system.

\item {\bf Broader impacts}
    \item[] Question: Does the paper discuss both potential positive societal impacts and negative societal impacts of the work performed?
    \item[] Answer: \answerNo{}.
    \item[] Justification: The paper discusses practical implications for downstream representation use in Appendix~\ref{sec:limitations}, but it does not include a separate broader-impact discussion covering both positive and negative societal impacts. The work is foundational and diagnostic rather than a deployed application or a release of a high-risk capability.
    
\item {\bf Safeguards}
    \item[] Question: Does the paper describe safeguards that have been put in place for responsible release of data or models that have a high risk for misuse (e.g., pre-trained language models, image generators, or scraped datasets)?
    \item[] Answer: \answerNA{}.
    \item[] Justification: The paper does not release a new high-risk model, image generator, language model, or scraped dataset. It analyzes self-supervised vision encoders using existing benchmark datasets and protocols.

\item {\bf Licenses for existing assets}
    \item[] Question: Are the creators or original owners of assets (e.g., code, data, models), used in the paper, properly credited and are the license and terms of use explicitly mentioned and properly respected?
    \item[] Answer: \answerYes{}.
    \item[] Justification: Appendix~\ref{app:asset_licenses} lists the datasets, benchmarks, codebases, and model releases used in the paper, together with their original credits, official sources, and licenses or terms of use. We follow these terms and do not redistribute raw third-party images or videos in the supplementary material unless permitted by the original asset provider.

\item {\bf New assets}
    \item[] Question: Are new assets introduced in the paper well documented and is the documentation provided alongside the assets?
    \item[] Answer: \answerNA{}.
    \item[] Justification: The paper introduces new annotation-free diagnostics and documents their protocols in Appendices~\ref{app:descriptor_decodability} and~\ref{app:local_global_details}, but it does not release a new dataset, model, or standalone asset in the current draft.

\item {\bf Crowdsourcing and research with human subjects}
    \item[] Question: For crowdsourcing experiments and research with human subjects, does the paper include the full text of instructions given to participants and screenshots, if applicable, as well as details about compensation (if any)? 
    \item[] Answer: \answerNA{}.
    \item[] Justification: The work does not involve crowdsourcing, user studies, or new research with human subjects.

\item {\bf Institutional review board (IRB) approvals or equivalent for research with human subjects}
    \item[] Question: Does the paper describe potential risks incurred by study participants, whether such risks were disclosed to the subjects, and whether Institutional Review Board (IRB) approvals (or an equivalent approval/review based on the requirements of your country or institution) were obtained?
    \item[] Answer: \answerNA{}.
    \item[] Justification: The paper does not conduct research with human subjects, so IRB approval or equivalent review is not applicable.

\item {\bf Declaration of LLM usage}
    \item[] Question: Does the paper describe the usage of LLMs if it is an important, original, or non-standard component of the core methods in this research? Note that if the LLM is used only for writing, editing, or formatting purposes and does \emph{not} impact the core methodology, scientific rigor, or originality of the research, declaration is not required.
    \item[] Answer: \answerNA{}.
    \item[] Justification: The core methodology and experiments do not use LLMs as an important, original, or non-standard component.

\end{enumerate}

\end{document}